\def\year{2022}\relax
\PassOptionsToPackage{table}{xcolor}
\documentclass[letterpaper]{article} 
\usepackage{aaai22}  
\usepackage{times}  
\usepackage{helvet}  
\usepackage{courier}  
\usepackage[hyphens]{url}  
\usepackage{graphicx} 
\def\UrlFont{\rm}  
\usepackage{natbib}  
\usepackage{caption} 
\DeclareCaptionStyle{ruled}{labelfont=normalfont,labelsep=colon,strut=off} 
\usepackage{algorithm}

\usepackage{newfloat}
\usepackage{listings}
\floatstyle{ruled}
\newfloat{listing}{tb}{lst}{}
\floatname{listing}{Listing}
\title{DAFNe: A One-Stage Anchor-Free Deep Model for Oriented Object Detection}

\author{
 Steven Lang,\textsuperscript{\rm 1} Fabrizio Ventola, \textsuperscript{\rm 1}
 Kristian Kersting \textsuperscript{\rm 1,}\textsuperscript{\rm 2} \\
} \affiliations{
 \textsuperscript{\rm 1} TU Darmstadt, Darmstadt, Germany \\
 \textsuperscript{\rm 2} Hessian Center for AI and Centre for Cognitive Science, Darmstadt, Germany \\
 \{steven.lang, ventola, kersting\}@cs.tu-darmstadt.de }

\usepackage{booktabs} \usepackage{tabularx}

\usepackage[subrefformat=parens,labelformat=parens]{subcaption}
 \usepackage{mathtools}
\usepackage[export]{adjustbox}
\usepackage{algpseudocode,amsfonts,algorithmicx}
\usepackage[hidelinks]{hyperref}  

\hypersetup{
    colorlinks,
    linkcolor={blue!50!black},
    citecolor={blue!50!black},
    urlcolor={blue!50!black}
}

\usepackage{tikz} \usetikzlibrary{svg.path} \usetikzlibrary{positioning}
\usetikzlibrary{calc}
\usepackage{multirow}

\newcommand{\mathbold}[1]{\boldsymbol{#1}}
\newcommand{\mbc}{\mathbold{c}}
\newcommand{\mbX}{\mathbold{X}}
\newcommand{\mbt}{\mathbold{t}}

\newcommand{\mstrain}{\textsuperscript{*}}

\newcommand{\rnS}{R-50}
\newcommand{\rnM}{R-101}

\newcommand{\hgM}{HG-104}
\newcommand{\dlaS}{DLA-34}

\newcommand{\darknetS}{DN-53}
\newcommand{\twostage}{\textsuperscript{\textdagger}}

\definecolor{rowhighlightcolor}{HTML}{ffecb3}

\newcommand{\midrulethin}{\specialrule{.025em}{.3em}{.3em}}

\usepackage{xargs}
\usepackage[colorinlistoftodos,textsize=small]{todonotes}
\newcommandx{\todoc}[2][1=]{{\todo[linecolor=orange,backgroundcolor=orange!25,bordercolor=orange,#1]{\tiny
      TODO: #2}}}
\newcommandx{\unsure}[2][1=]{{\todo[linecolor=red,backgroundcolor=red!25,bordercolor=red,#1]{\tiny
      UNSURE: #2}}}
\newcommandx{\change}[2][1=]{{\todo[linecolor=blue,backgroundcolor=blue!25,bordercolor=blue,#1]{\tiny
      CHANGE: #2}}}
\newcommandx{\info}[2][1=]{{\todo[linecolor=green,backgroundcolor=green!25,bordercolor=green,#1]{\tiny
      INFO: #2}}}
\newcommandx{\improvement}[2][1=]{{\todo[linecolor=violet,backgroundcolor=violet!25,bordercolor=violet,#1]{\tiny
      IMPROVEMENT: #2}}}
\newcommandx{\thiswillnotshow}[2][1=]{{\todo[disable,#1]{THIS WILL NOT SHOW:
      #2}}}

\usepackage{gensymb} \usepackage[utf8]{inputenc} \usepackage{pgfplots}
\DeclareUnicodeCharacter{2212}{−} \usepgfplotslibrary{groupplots,dateplot}
\usetikzlibrary{patterns,shapes.arrows} \pgfplotsset{compat=newest}
\usepackage{layouts}

\begin{document}

\title{DAFNe: A One-Stage Anchor-Free Approach for Oriented Object
  Detection} 

\maketitle

\begin{abstract}
  We present DAFNe, a Dense one-stage Anchor-Free deep Network for oriented
  object detection. As a one-stage model, it performs bounding box
  predictions on a dense grid over the input image, being architecturally
  simpler in design, as well as easier to optimize than its two-stage
  counterparts. Furthermore, as an anchor-free model, it reduces the prediction
  complexity by refraining from employing bounding box anchors. With DAFNe we introduce an
  orientation-aware generalization of the center-ness function for arbitrarily
  oriented bounding boxes to down-weight low-quality predictions and a
  center-to-corner bounding box prediction strategy that improves object
  localization performance. Our experiments show that DAFNe outperforms all
  previous one-stage anchor-free models on DOTA 1.0, DOTA 1.5, and UCAS-AOD and
  is on par with the best models on HRSC2016.


\end{abstract}

\section{Introduction}
\label{sec:introduction}

\begin{figure}
  \begin{center}
    \includegraphics[width=0.24\linewidth,height=0.24\linewidth]{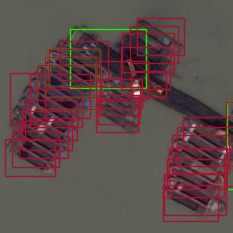}\hspace{0.1em}%
    \includegraphics[width=0.24\linewidth,height=0.24\linewidth]{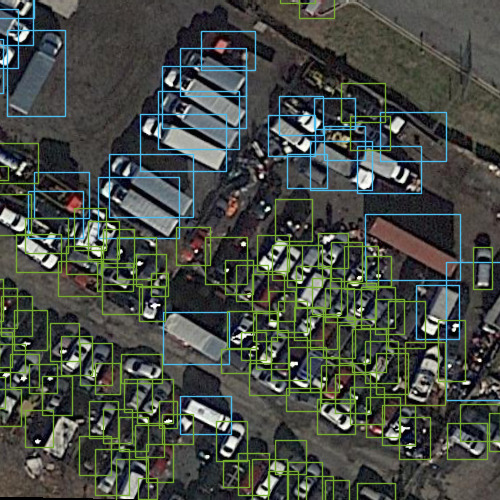}\hspace{0.1em}%
    \includegraphics[width=0.24\linewidth,height=0.24\linewidth]{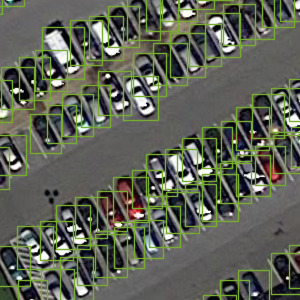}\hspace{0.1em}%
    \includegraphics[width=0.24\linewidth,height=0.24\linewidth]{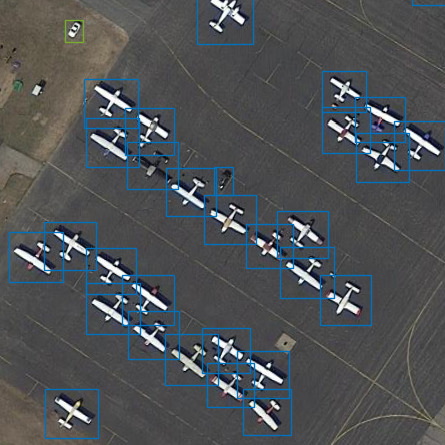}\hspace{0.1em}
    \includegraphics[width=0.24\linewidth,height=0.24\linewidth]{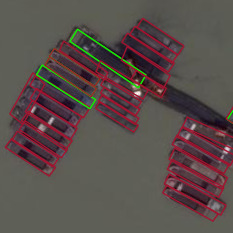}\hspace{0.1em}%
    \includegraphics[width=0.24\linewidth,height=0.24\linewidth]{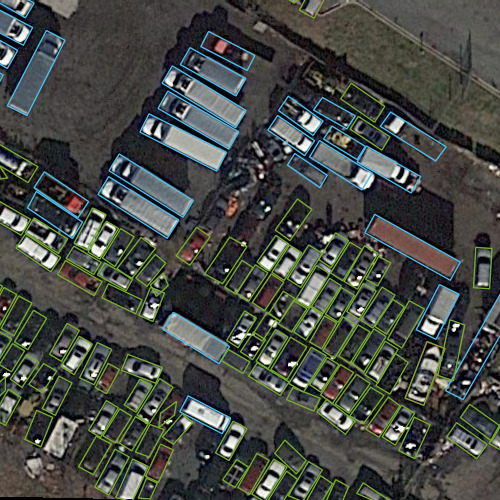}\hspace{0.1em}%
    \includegraphics[width=0.24\linewidth,height=0.24\linewidth]{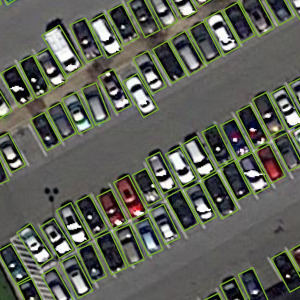}\hspace{0.1em}%
    \includegraphics[width=0.24\linewidth,height=0.24\linewidth]{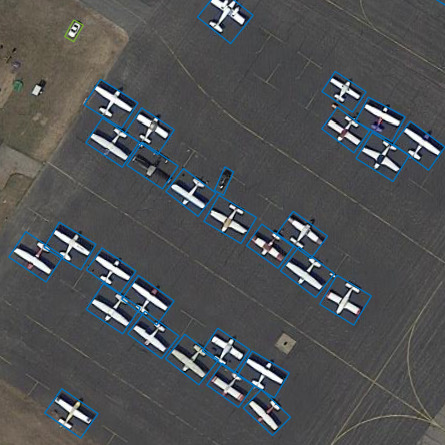}\hspace{0.1em}
  \end{center}
  \caption{ Comparison of axis-aligned (top) and oriented (bottom) bounding box
    detections obtained with our model on aerial view imagery from DOTA 1.0.
    Clearly, the oriented bounding box representation is a generalization of
    horizontal bounding boxes and therefore a better choice for oriented
    objects. It generates tighter boxes that better cover and fit the true area
    of the object.}
  \label{fig:rot-box-vs-horiz-box}
\end{figure}

Object detection is one of the fundamental tasks in computer vision that is
needed to develop intelligent systems which aim to understand the visual world.
The task of object detection is to answer two main questions: \textit{Where} are
visual objects of interest located in a given image (object localization)?
\textit{What} are these objects (object recognition)? Both questions are tightly
coupled as we first need to understand where to look before asking what we are
looking at. Many computer vision problems depend on object detection as a basis:
Object tracking~\cite{8003302} performs object detection in a sequence of images
and matches the corresponding objects across time, instance
segmentation~\cite{instance-aware-semantiv-seg,simul-det-and-seg,hypercolumns,mask-rcnn}
predicts labels of different object instances pixel-wise, image
captioning~\cite{karpathy2015deep,Wu2018ImageCA,pmlr-v37-xuc15} generates a
textual description of the image content. Recently, a generalization of object
detection has emerged: oriented object detection. Here, the localization extends
to arbitrarily oriented objects by taking the object orientation into account.
As demonstrated in \figurename~\ref{fig:rot-box-vs-horiz-box}, oriented bounding
boxes are a generalization of horizontal bounding boxes and can tightly fit
objects with dense object distributions and arbitrary orientations.

As with several tasks in machine learning, the rise of deep
learning~\cite{goodfellow2016deeplearning} in recent years has pushed the
boundaries in computer vision far beyond what was prior state-of-the-art. Most
successful deep architectures for object detection such as Faster
R-CNN~\cite{faster-rcnn}, SSD~\cite{ssd}, and YOLOv2,v3~\cite{yolo-v2,yolo-v3}
are based on a technique that includes prior information about possible object
sizes and aspect ratios. These are called bounding box priors or \textit{anchor
  boxes} which need to be handcrafted and can vary for different datasets,
introducing many additional hyper-parameters to be tuned for a single dataset.
On the other side, humans can localize objects without pre-defined templates.
Motivated by this idea, anchor-free approaches forgo prior information about
possible bounding boxes and, thus, they get rid of the additional
hyper-parameters and reduce the computational complexity. This is the direction
taken by UnitBox~\cite{unitbox}, YOLOv1~\cite{yolo}, CenterNet~\cite{centernet},
CornerNet~\cite{cornernet}, ExtremeNet~\cite{extremenet},
FoveaBox~\cite{kong2019foveabox}, and FCOS~\cite{fcos}, to name a few. Moreover,
object detection approaches can be divided into two main categories. The first
includes those methods that are based on a \textit{two-stage} architecture,
where the first stage generates regions of interest proposals where possible
objects could be located, while the second stage takes each proposal separately,
fine-tunes the localization, and classifies the object. The second category
consists of \textit{one-stage} methods such as RetinaNet~\cite{retinanet},
SSD~\cite{ssd}, YOLOv1,v2,v3~\cite{yolo,yolo-v2,yolo-v3}, which get rid of the
region proposal stage and perform the localization and classification directly
on a dense grid of possible locations in a single pass. As shown in
RetinaNet~\cite{retinanet}, these can achieve competitive accuracy while being
faster than their two-stage alternatives.

Oriented object detection has been approached mainly by two-stage anchor-based
models in the past, while only a few works have focused on the faster and
conceptually simpler one-stage anchor-free models. In this work, our goal is to
approach oriented object detection with a one-stage anchor-free deep model,
taking advantage of their simpler design and reduced computational requirements.
To this aim, inspired by the success of prominent one-stage anchor-free models
from horizontal object detection, we introduce DAFNe, a novel dense one-stage
anchor-free approach for oriented object detection. Moreover, we propose an
orientation-aware generalization of the center-ness function presented in
FCOS~\cite{fcos} to focus on predictions that are closer to an oriented object's
center. We also enhance the detector's localization performance by introducing a
novel center-to-corner bounding box regression to reframe the corner regression
by separating it into an easier object-center prediction and a subsequent
center-to-corner prediction step. These contributions can improve the model's
detection accuracy outperforming, to the best of our knowledge, all one-stage
anchor-free models in the literature on the most common benchmarks.
To summarize, our contributions are the
following:
\begin{itemize}
  \item We introduce DAFNe, a novel one-stage anchor-free deep model for
        oriented object detection that, to date, is the most accurate
        one-stage anchor-free architecture for this task.
  \item We present the first generalization of the center-ness function to
        arbitrary quadrilaterals that take into account the object's
        orientation and that accurately down-weights low-quality
        predictions.
  \item We reframe the corner regression following a divide-and-conquer
        strategy, i.e. we divide it into an easier object-center prediction step
        followed by a center-to-corner prediction to improve object localization
        performance.
\end{itemize}

\section{Related Work}
\label{sec:background}

Due to their success in horizontal object detection, the vast majority of the
work on oriented object detection has focused on two-stage anchor-based
architectures~\cite{roi-trans,Chen2017RethinkingAC,Li_2019_CVPR_Workshops,Liu2017LearningAR,Jianqi17RRPN,yang2020scrdet}.
Later contributions have shifted from two-stage to one-stage models, taking
advantage of their architectural and procedural simplicity. Here we briefly
introduce the most relevant anchor-based and anchor-free one-stage models in the
literature.

\subsection{One-Stage Oriented Object Detection: Anchor-Based Methods}
The first relevant work on one-stage anchor-based oriented object detection
introduces the Salience Biased Loss~\cite{sun2018salience}, which is a loss
function based on salience information directly extracted from the input image.
This is similar in spirit to the Focal Loss in RetinaNet~\cite{retinanet}, as it
treats training samples differently according to their complexity (saliency).
This is estimated by an additional deep model trained on
ImageNet~\cite{imagenet_cvpr09} where the number of active neurons across
different convolution layers is measured. The concept is that with increasing
complexity, more neurons would be active. The saliency then scales an arbitrary
base loss function to adapt the importance of training samples accordingly.

The work by Han et al.~\cite{han2020align} (S2ANet) attempts to improve the discrepancy
of classification score and localization accuracy and ascribes this issue to the
misalignment between anchor boxes and the axis-aligned convolutional features.
Hence, the authors propose two modules. The first one generates high
quality-oriented anchors using their Anchor Refinement Network which adaptively
aligns the convolutional features according to the generated anchors using an
Alignment Convolution Layer. The second module adopts Active Rotating
Filters~\cite{orn} to encode the orientation context and it produces
orientation-aware features to alleviate classification score and localization
accuracy inconsistencies.

In R\textsuperscript{3}Det~\cite{Yang2021R3DetRS}, the authors propose a progressive
regression approach from coarse to fine granularity, introducing a feature
refinement module that re-encodes the current refined bounding box to the
corresponding feature points through pixel-wise feature interpolation, realizing
feature reconstruction and alignment.

Besides, the solution devised by Yang et al.\cite{yang2020arbitrary} tackles the
issue of discontinuous boundary effects on the loss due to the inherent angular
periodicity and corner ordering by transforming the angular prediction task from
a regression problem into a classification problem. The authors conceived the
Circular Smooth Label technique which handles the periodicity of angles and
raises the error lenience to adjacent angles.

\subsection{One-Stage Oriented Object Detection: Anchor-Free Methods}

The first one-stage anchor-free oriented object detector,
IENet~\cite{lin2019ienet}, is based on the one-stage anchor-free fully
convolutional detector FCOS. The regression head from FCOS is extended by
another branch that regresses the bounding box orientation, using a
self-attention mechanism that incorporates the branch feature maps of the object
classification and box regression branches.

In Axis-Learning~\cite{xiao2020axis}, the authors also build on the dense
sampling approach of FCOS and explore the prediction of an object axis, defined
by its head point and tail point of the object along its elongated side (which
can lead to ambiguity for near-square objects). The axis is extended by a width
prediction which is interpreted to be orthogonal to the object axis. Similar to
our work, Xiao et al.~\cite{xiao2020axis} further introduce an orientation
adjusted formulation for rectangular bounding boxes of the center-ness function,
proposed by FCOS, which is scaled by the bounding box aspect ratio to take
objects with heavily skewed aspect ratios into account.

In PIoU~\cite{chen2020piou} the authors argue that a distance-based regression
loss such as SmoothL1 only loosely correlates to the actual Intersection over
Union (IoU) measurement, especially in the case of large aspect ratios.
Therefore, they propose the Pixels-IoU (PIoU) loss, which exploits the IoU for
optimization by pixel-wise sampling, dramatically improving detection
performance on objects with large aspect ratios.

P-RSDet~\cite{zhou2020objects} replaces the Cartesian coordinate representation
of bounding boxes with polar coordinates. Therefore, the bounding box regression
happens by predicting the object's center point, a polar radius, and two polar
angles. Furthermore, they introduce the Polar Ring Area Loss to express the
geometric constraint between the polar radius and the polar angles.

Another alternative formulation of the bounding box representation is defined in
O\textsuperscript{2}-DNet~\cite{wei2020o2-dnet}. Here, oriented objects are
detected by predicting a pair of middle lines inside each target, similar to the
extreme keypoint detection schema proposed in ExtremeNet~\cite{extremenet}.

\section{DAFNe: A One-Stage Anchor-Free Approach for Oriented Object
  Detection}\label{sec:method}

Oriented object detection has been mainly tackled by the adaption of two-stage
anchor-based models, which come with increased methodological and computational
complexity. In this work, we take a different path and propose DAFNe, an
oriented object detector, devising a one-stage anchor-free architecture that is
simpler and faster in concept compared to its two-stage anchor-based
alternatives. We employ RetinaNet~\cite{retinanet} as the base architecture,
which is composed of a ResNet~\cite{resnet} backbone combined with a Feature
Pyramid Network (FPN)~\cite{fpns} as multi-scale feature extractor on top which
produces feature maps at five scales (see Appendix A for the architecture
schematic). The different feature maps are then fed into a prediction head,
consisting of two sequential convolution branches for classification and
regression respectively. To regress oriented bounding boxes, we construct
relative targets such that
\begin{align}
  \mbt^{*} \leftarrow ( & x_{0}^* - x, y_{0}^* - y, x_{1}^* - x, y_{1}^* - y, \nonumber \\
                        & x_{2}^* - x, y_{2}^* - y, x_{3}^* - x, y_{3}^* - y) / s \, ,
\end{align}
where $(x_{i}^{*},y_{i}^{*})$ are the absolute corner coordinates of the
original target bounding box, $(x, y)$ is a location in the regression branch
feature map and $s=2^{l}$ is the output stride of level $l$ of the FPN feature
map. Thus, the target bounding boxes are then represented as scale-normalized
offsets from locations w.r.t. the output feature map. For further architectural,
training, and inference details we refer the reader to
RetinaNet~\cite{retinanet} and FCOS~\cite{fcos}.

\subsection{Oriented Center-Ness}
\label{sec:method:oriented-center-ness}

\begin{figure}
  \centering
  \begin{subfigure}{0.5\linewidth}
    \centering \includegraphics{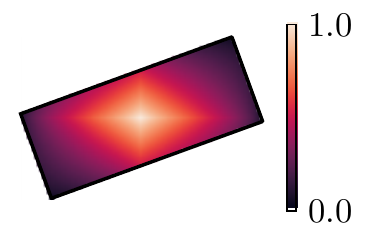} \subcaption{Center-ness}
    \label{fig:method:oriented-centerness-main:center-ness}
  \end{subfigure}%
  \begin{subfigure}{0.5\linewidth}
    \centering \includegraphics{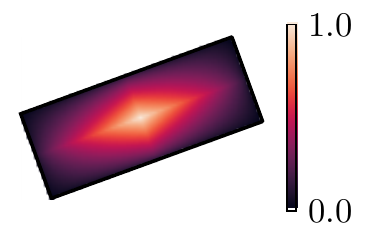}
    \subcaption{Oriented Center-ness}
    \label{fig:method:oriented-centerness-main:oriented-center-ness}
  \end{subfigure}
  \caption{Function evaluation of every location in a given oriented bounding
    box w.r.t. the original
    center-ness~\subref{fig:method:oriented-centerness-main:center-ness} and the
    proposed oriented
    center-ness~\subref{fig:method:oriented-centerness-main:oriented-center-ness}.
    It is clear that the original center-ness function does not adapt well to
    oriented bounding boxes. On the contrary, the proposed oriented center-ness
    aligns to arbitrary edge orientations and it can take heavily skewed aspect
    ratios into account as well.}
  \label{fig:method:oriented-centerness-main}
\end{figure}

FCOS~\cite{fcos} has introduced the \textit{center-ness} prediction which
depicts the normalized distance from the feature map location to the center of
the target object, for which the location is responsible. Since center-ness has
been formulated in the task of classic horizontal object detection and it is
measured w.r.t. axis-aligned edges, it does not directly translate to oriented
bounding boxes when applied to their axis-aligned hull (see
\figurename~\ref{fig:method:oriented-centerness-main:center-ness}).

Therefore, we propose a generalization of the center-ness formulation to
arbitrary quadrilaterals by measuring the normalized perpendicular distances
between any feature map location $(x, y)$ and the edge between two subsequent
corners $(p^{0}_{x}, p^{0}_{y})$ and $(p^{1}_{x}, p^{1}_{y})$ in the oriented
bounding box:
\begin{align}
  &\text{dist} \left( p^{0}_{x}, p^{0}_{y}, p^{1}_{x}, p^{1}_{y},  x, y \right)
  \nonumber \\
  = &\frac{\left| \left(p^{1}_{x} - p^{0}_{x}\right) \left(p^{0}_{y} - y\right) - \left( p^{0}_{x} - x \right) \left( p^{1}_{y} - p^{0}_{y} \right) \right|}
    {\sqrt{ \left( p^{1}_{x} - p^{0}_{x} \right)^{2} + \left( p^{1}_{y} - p^{0}_{y} \right)^{2}}} \, .
\end{align}
With this, we can obtain the four perpendicular distances $a, b, c, d$ of a
given oriented bounding box $(x_{0}$, $y_{0}$, $x_{1}$, $y_{1}$, $x_{2}$,
$y_{2}$, $x_{3}$, $y_{3})$:
\begin{align}
  a & = \text{dist} \left( x_{0}, y_{0}, x_{1}, y_{1} , x, y \right), \nonumber  \\
  b & = \text{dist} \left( x_{1}, y_{1}, x_{2}, y_{2} , x, y \right)    \nonumber  \\
  c & = \text{dist} \left( x_{2}, y_{2}, x_{3}, y_{3} , x, y \right), \nonumber \\
  d & = \text{dist} \left( x_{3}, y_{3}, x_{0}, y_{0} , x, y \right) \, . \label{eq:abcd}
\end{align}
The oriented center-ness is then calculated with
\begin{align}
  \text{center-ness} \left( a, b, c, d \right) = \left(\frac{\text{min} \left( a, c \right)}{\text{max} \left( a, c \right)} \cdot \frac{\text{min} \left( b, d \right)}{\text{max} \left( b, d \right)} \right)^{\frac{1}{\alpha}}  \, ,
  \label{eq:method:center-ness}
\end{align}
where the hyper-parameter $\alpha$ controls the decay rate of the center-ness.
\figurename~\ref{fig:method:oriented-centerness-main:oriented-center-ness} shows
the oriented center-ness heatmap for each location in the given oriented
bounding box. One can see that the function now aligns perfectly with the box
edges and adapts to its aspect ratio whereas the original center-ness formulation
depicted in \figurename~\ref{fig:method:oriented-centerness-main:center-ness}
fails to do so. The center-ness value ranges from $0$ to $1$, corresponding to
the location being on the edge or in the center of the box
respectively. 

The goal of oriented center-ness is to down-weight classification scores of
predicted low-quality bounding boxes and, thus, removing them during
post-processing steps such as confidence thresholding and non-maximum
suppression. In this context, a low-quality bounding box is defined as a
prediction with high confidence but low overlap with the ground-truth bounding
box which possibly leads to false positives and, consequently, reduces the
detectors precision~\cite{fcos}.

The oriented center-ness is predicted using an additional $3 \times 3$
convolution with a single output channel on top of the regression branch. During
testing, the final score $s_{x, y}$, used for ranking detections during
non-maximum suppression is obtained by adjusting the classification confidence
$p_{x, y}$ with the predicted oriented center-ness $o_{x, y}$ at location
$(x, y)$: $s_{x, y} = \sqrt{p_{x, y} \cdot o_{x, y}}$.

Another modification of the center-ness function to oriented bounding boxes has
been proposed in Axis-Learning~\cite{xiao2020axis}. It is limited to oriented
rectangles and not normalized, i.e. depending on the object aspect ratio, it can
exceed values of $1$. In contrast, our formulation generalizes to arbitrary
quadrilaterals and it is constrained to be in $\left[0, 1\right]$.

\subsection{Corner Prediction Strategies}
\label{sec:method:corner-prediction-strategies}
\begin{figure*}
  \begin{subfigure}{0.250\linewidth}
    \includegraphics[width=1.0\textwidth]{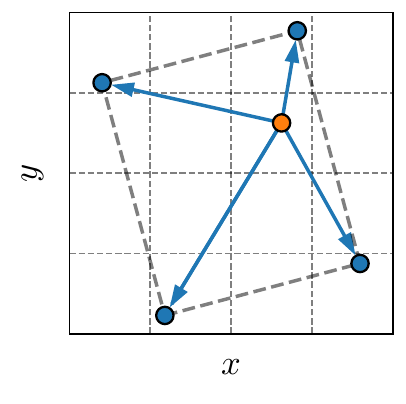}
    \subcaption{Direct}
    \label{fig:method:corner-prediction-strategies:direct}
  \end{subfigure}%
  \begin{subfigure}{0.250\linewidth}
    \includegraphics[width=1.0\textwidth]{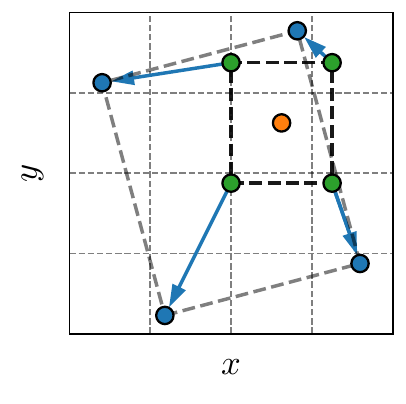}
    \subcaption{Offset}
    \label{fig:method:corner-prediction-strategies:anchor-offset}
  \end{subfigure}%
  \begin{subfigure}{0.250\linewidth}
    \includegraphics[width=1.0\textwidth]{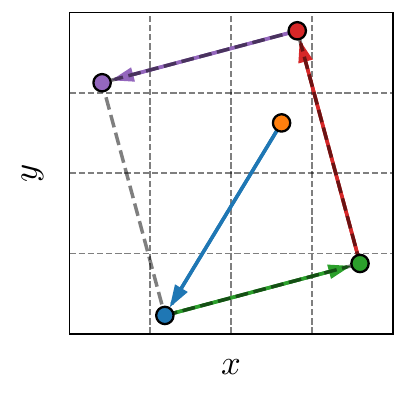}
    \subcaption{Iterative}
    \label{fig:method:corner-prediction-strategies:iterative}
  \end{subfigure}%
  \begin{subfigure}{0.250\linewidth}
    \includegraphics[width=1.0\textwidth]{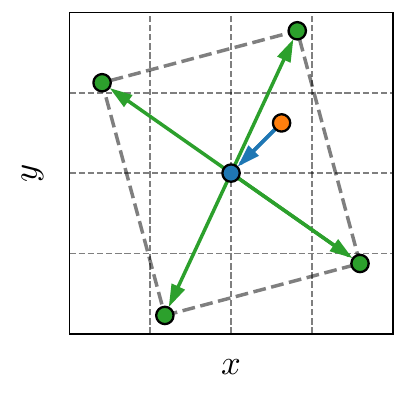}
    \subcaption{Center-to-Corner}
    \label{fig:method:corner-prediction-strategies:center-to-corner}
  \end{subfigure}
  \caption{The baseline \subref{fig:method:corner-prediction-strategies:direct} and the
    proposed \subref{fig:method:corner-prediction-strategies:anchor-offset}, \subref{fig:method:corner-prediction-strategies:iterative},
    and \subref{fig:method:corner-prediction-strategies:center-to-corner} corner
    prediction strategies. In
    \subref{fig:method:corner-prediction-strategies:direct}, the four corners
    are obtained by predicting four vectors as an offset from the current
    location. \subref{fig:method:corner-prediction-strategies:anchor-offset}
    resembles an anchor-based approach with a single base anchor, it predicts
    the corners as offsets for each corner of the anchor. In
    \subref{fig:method:corner-prediction-strategies:iterative}, each corners is
    obtained by taking the previous corner into consideration. In
    \subref{fig:method:corner-prediction-strategies:center-to-corner}, first the
    center is predicted as an offset from the location, then, the corners are
    predicted as an offset from the predicted center.}
  \label{fig:method:corner-prediction-strategies}
\end{figure*}

\noindent Traditionally, in horizontal and oriented object detection, the bounding box
regression is performed in a single step, i.e. the last feature map of the
regression branch predicts all coordinates at a specific feature location at
once. This is shown in
\figurename~\ref{fig:method:corner-prediction-strategies:direct} where each
corner is predicted as an offset from the location (orange dot). In this work,
we explore the idea of separating the prediction of corners into multiple steps,
by introducing direct dependencies between the intermediate steps. To the best
of our knowledge, no previous work has explored and analyzed the impact of
dividing the bounding box prediction into multiple interdependent parts. We
compare the baseline of ``direct'' box predictions to a naive
single-anchor-based approach named ``offset'' depicted in
\figurename~\ref{fig:method:corner-prediction-strategies:anchor-offset}, as well
as to two prediction strategies we introduce, i.e. ``iterative'' and
``center-to-corner''.

The ``iterative'' prediction strategy is inspired by the decomposition of a
joint probability distribution into conditionals using the chain rule. This
leads to an iterative prediction strategy where each corner is predicted using
the previous corners as shown in
\figurename~\ref{fig:method:corner-prediction-strategies:iterative}:
\begin{align}
  \mbc_{0} & = \text{conv}_{256 \rightarrow 2} \left( \mbX \right)                               \nonumber     \\
  \mbc_{1} & = \text{conv}_{258 \rightarrow 2} \left( \mbX, \mbc_{0} \right)                      \nonumber    \\
  \mbc_{2} & = \text{conv}_{260 \rightarrow 2} \left( \mbX, \mbc_{0}, \mbc_{1} \right)           \nonumber     \\
  \mbc_{3} & = \text{conv}_{262 \rightarrow 2} \left( \mbX, \mbc_{0}, \mbc_{1}, \mbc_{2} \right) \, , \label{eq:iterative-corner-pred}
\end{align}
where $\text{conv}_{ic \rightarrow oc}$ represents a $3 \times 3$ convolution
with $ic$ input channels and $oc$ output channels. The idea is that with more
information available, following corners can be predicted more consistently
w.r.t. the previous ones. As an extreme example, knowing the first three corners
and predicting the fourth from those should be easier than predicting the fourth
corner without prior knowledge about where the other corners are (under the
assumption that this is not already encoded in the incoming features which are
used to predict the others).

The second proposal, the ``center-to-corner'' strategy, is to split the
prediction into two phases. The first phase predicts the center of the object
while the second phase regresses the corners from the predicted center (see
\figurename~\ref{fig:method:corner-prediction-strategies:center-to-corner}). For
this, the regression branch is divided into two separate branches, namely the
center branch responsible for the object's center prediction and the
corner branch responsible for the center-to-corner offset prediction.
\begin{align}
  \tilde{\mbc}                           & = \text{center-branch}_{256 \rightarrow 2} \left( \mbX \right) \nonumber                                                              \\
  \mbc_{0}, \mbc_{1}, \mbc_{2}, \mbc_{3} & = \text{corner-branch}_{256 \rightarrow \left( 4 \times 2 \right)} \left(\mbX \right) + \tilde{\mbc} \, . \label{eq:center-to-corner-pred}
\end{align}
Note that $+$ in \eqref{eq:center-to-corner-pred} is a broadcasting operation
and adds the center prediction $\tilde{\mbc}$ to each tuple of the corner branch
output. In addition, the center prediction is also used during the optimization,
such that the center is not only indirectly optimized using the corners as a
proxy but also directly, by employing a loss w.r.t. the target object center.
This strategy is inspired by the general concept of divide-and-conquer. In fact,
the task of predicting the center is easier than the task of predicting all four
corners as it is lower-dimensional and is independent of the object's width,
height, and orientation. Therefore, the model can first focus on getting the
general location of an object right by optimizing the center loss and, then,
gradually refine the corners from the predicted center.

\subsection{Learning Objectives}
\label{sec:method:learning-objectives}

We follow RetinaNet~\cite{retinanet} for the choice of learning objectives. That
is, the Focal Loss is employed for classification to down-weight the
contribution of objects that are easy to classify and shift the training focus
on objects that are predicted with low confidences. For the regression of
oriented bounding box coordinates, we adopt the SmoothL1 loss over the
difference between the predicted and the target coordinates. To increase
training stability, we use the eight-point loss modulation proposed by Qian et
al.~\cite{qian2019learning}. As in FCOS, the oriented center-ness is interpreted
as the likelihood of the responsibility of the current location w.r.t. the
target. Thus, the binary cross-entropy loss is used as the oriented center-ness
optimization objective.

\section{Experimental Evaluation}
\label{sec:exp}

In this section, we empirically evaluate our contributions and show how they can
be beneficial for oriented object detection. We run experiments to answer the
following questions: (\textbf{Q1}) Does the proposed oriented center-ness
improve accuracy? (\textbf{Q2}) Do the introduced corner prediction strategies
perform better than the approach commonly used in the literature? (\textbf{Q3})
Does DAFNe, together with oriented center-ness and center-to-corner prediction
strategy, improve the state-of-the-art accuracy for one-stage anchor-free
oriented object detection?

\subsection{Experimental Setup}
\label{sec:exp:experimental-setup}

\paragraph{Dataset.}
\label{sec:exp:dataset}
The experiments are conducted on HRSC2016~\cite{hrsc2016},
UCAS-AOD~\cite{li2019ucas-aod}\footnote{For the UCAS-AOD benchmark we use the predefined split from \url{https://github.com/ming71/UCAS-AOD-benchmark} to ensure comparability between results.}, DOTA 1.0~\cite{Xia_2018_CVPR}, and DOTA 1.5
datasets. For DOTA 1.0 and 1.5 we pre-process each image into patches of
$1024 \times 1024$ pixels with an overlap of $200$ pixels between patches.
Online data augmentation is applied during training. Specifically, we perform
random horizontal and vertical flipping with a chance of $50$\%. Additionally,
the images are rotated by either $0\degree$, $90\degree$, $180\degree$ or
$270\degree$ with equal chance. As commonly done in practice, for our final
results (Table~\ref{tab:exp:comparison}) we train the model on the joint of
train and validation set, increase the number of training iterations by a factor
of 3, train on image patches of multiple patch-sizes (600, 800, 1024, 1300,
1600) on DOTA, and adopt multi-scale train and test-time augmentation (450, 500,
600, 700, 800, 900, 1000, 1100, 1200). As an accuracy measurement, we adopt the
mean Average Precision (mAP) evaluation metric.

\paragraph{Training and Inference.}
We employ ResNet-50 as the backbone model unless otherwise specified. The
backbone weights are initialized from a pre-trained ResNet-50 on ImageNet. All
other model weights are initialized as in RetinaNet~\cite{retinanet}. The model
is optimized using SGD for $90$k iterations with a base learning rate of $0.01$
and a batch size of $8$ images. The learning rate is reduced by a factor of $10$
after $60$k and $80$k iterations. Inspired by Goyal et
al.~\cite{goyal2017accurateLM}, we employ a warm-up phase of $500$ iterations
with linear scaling of the learning rate starting from $0.001$ to $0.01$ to
overcome optimization difficulties in early training. The weight-decay and
momentum optimization techniques are applied with values of $0.0001$ and $0.9$
respectively. The loss function hyper-parameters were adapted from RetinaNet
without further tuning, i.e. for Focal Loss we set $\alpha=0.25$ and
$\gamma=2.0$, while for SmoothL1 loss we set $\beta = 1/9$. To account for the
magnitude discrepancy between the classification and regression loss, the
classification loss weight is set to $\lambda_{cls} = 10$ and for all other loss
terms we set $\lambda_{i} = 1$. Additionally, all loss weights were normalized
setting $\lambda_{i} = \lambda_{i} / \sum_{j} \lambda_{j}$. At test time,
predictions with a confidence $p < 0.05$ are removed. Furthermore, we apply a
\textit{top-k} class-confidence filtering with $k=2000$. For non-maximum
suppression, we set the threshold $t_{\text{nms}}$ for overlapping boxes of the
same class to $0.1$. All experiments were run on NVIDIA DGX-2 systems, using
four V100 (32GB VRAM) GPUs.

\subsection{(Q1) Oriented Center-Ness Improves Accuracy}
\label{sec:exp:oriented-center-ness}

\begin{figure*}[ht!]
  \centering
  \begin{subfigure}{0.5\linewidth}
    \centering
    \includegraphics[width=1.0\linewidth]{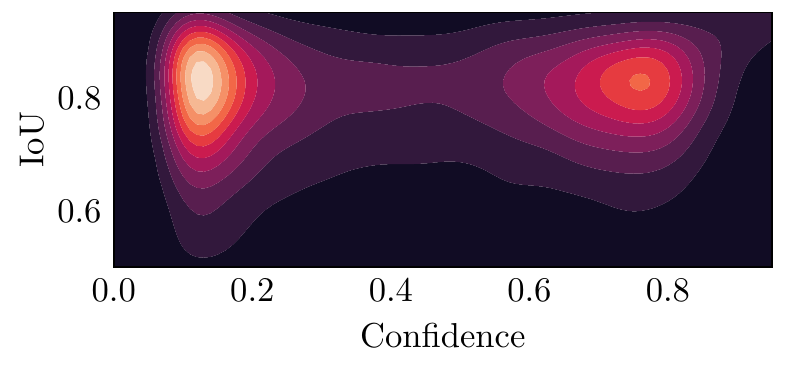}%
    \caption{Without oriented center-ness}
    \label{fig:exp:oriented-center-ness:without}
  \end{subfigure}%
  \begin{subfigure}{0.5\linewidth}
    \centering
    \includegraphics[width=1.0\linewidth]{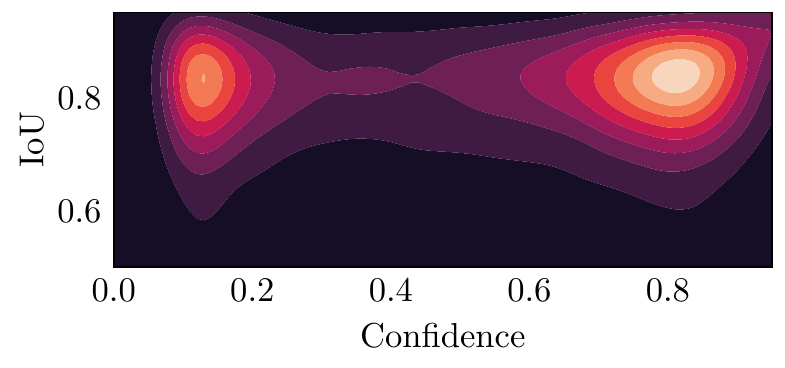}
    \caption{With oriented center-ness}
    \label{fig:exp:oriented-center-ness:with}
  \end{subfigure}%

  \caption{Density of classification confidences against IoU values for
    correctly detected (true positives) oriented bounding boxes on DOTA 1.0 validation set.
    Predictions adjusted with oriented center-ness lead to higher
    classification confidence. As a result, accuracy is improved since
    detections with low center-ness values are successfully removed during
    post-processing steps such as confidence thresholding and non-maximum
    suppression. }
  \label{fig:exp:oriented-center-ness}
\end{figure*}

\begin{table*}
  \caption{Influence of the center-ness function on accuracy (mAP) on the
    validation set of HRSC2016, UCAS-AOD, DOTA 1.0, and DOTA 1.5. The introduced oriented
    center-ness function consistently improves model accuracy on all datasets}
  \label{tab:exp:center-ness}
  \begin{center}
    \begin{tabular}{llrrrr}
      \toprule
      & Center-ness         & \phantom{x} HRSC2016  & \phantom{x} UCAS-AOD         & \phantom{x} DOTA 1.0       & \phantom{x} DOTA 1.5       \\
      \midrule
      \multirow{2}{4em}{Baseline} & none         & 61.68 & 89.83         & 69.97             & 68.91          \\
                                & axis-aligned & 70.30 &   89.97       & 69.95          & 67.98          \\
      \midrulethin
      Ours & oriented    & \textbf{78.87} & \textbf{90.03} & \textbf{71.17} & \textbf{70.09} \\
      \bottomrule
    \end{tabular}
  \end{center}
\end{table*}

\noindent We compare DAFNe with the proposed oriented center-ness against DAFNe without
center-ness and against DAFNe with axis-aligned center-ness (by using the
axis-aligned hull of the oriented bounding box). For this comparison, we employ
the direct corner prediction strategies on all models.
Table~\ref{tab:exp:center-ness} shows that the oriented center-ness consistently
improves the model's performance on all datasets. We observed that different
values for $\alpha$ are optimal for different datasets ($\alpha=4,3,5,2$ for DOTA
1.0, DOTA 1.5, UCAS-AOD, and HRSC2016 respectively). A larger value of $\alpha$ leads to a
slower decay of the oriented center-ness value, starting from $1$ in the center
of the object to $0$ at the object's boundaries. Since the predicted center-ness
is used to adjust the classification confidence value, $\alpha$ effectively controls
the penalty that off-center predictions receive. Thus, the proposed oriented
center-ness improves the validation mAP score by assigning low scores to
off-center predictions that are removed afterward, during post-processing steps.

To better understand the improvement that the oriented center-ness adjustment
to the classification confidence achieves, we have collected the true positive
detections on DOTA 1.0 as a heatmap in \figurename~\ref{fig:exp:oriented-center-ness} (see
Appendix B for a class-based separation of this heatmap). Here, the
classification confidences of the detections are drawn against the IoU with the ground-truth
bounding box. That is, given a trained model, for each correctly detected
oriented bounding box in the validation set, a pair of (\textit{confidence},
\textit{IoU}) is collected, where the IoU is computed between the predicted
bounding box and the ground-truth bounding box. The heatmap brightness indicates
the density of correctly detected objects in the validation dataset. The
classification confidences in the heatmap of
\figurename~\ref{fig:exp:oriented-center-ness:without} are obtained without
employing oriented center-ness, while confidences in the heatmap of
\figurename~\ref{fig:exp:oriented-center-ness:with} are obtained by employing
oriented center-ness (note that the values used for the heatmaps are the
confidences $p$ and not the center-ness adjusted scores $s$).
The inclusion of oriented center-ness as an
adjustment to the classification confidence results in detections of higher
confidences. Consequently, the detections with low center-ness values are
successfully removed during post-processing steps such as score thresholding and
non-maximum suppression. Thus, we can answer (\textbf{Q1}) affirmatively, the
proposed oriented center-ness is beneficial for the model accuracy.

\subsection{(Q2) Corner Prediction Strategies to Improve Detection}
\label{sec:exp:corner-prediction-strategies}

\begin{table*}
  \caption{Comparison of the proposed divide-and-conquer corner prediction
    strategies with the direct strategy, i.e. the de facto standard approach.
    Results are obtained on the validation sets. Center-to-corner is overall the
    best strategy. It outperforms the baseline (direct) strategy on every dataset (noted
    with $\uparrow$) and it outperforms the other novel methods we introduced
    except on DOTA 1.5, where it is on par with them}
  \label{tab:exp:prediction-strategies}
  \begin{center}
    \begin{tabular}{llrrrr}
      \toprule
      & Strategy         & \phantom{x} HRSC2016           & \phantom{x} UCAS-AOD & \phantom{x} DOTA 1.0       & \phantom{x} DOTA 1.5       \\
      \midrule
      \multirow{1}{4em}{Baseline} & direct           & 61.68  &   89.83        & 69.97          & 68.91          \\
      \midrulethin
       \multirow{2}{4em}{Ours}    & offset           & 64.84  &   89.95      & 71.10          & \textbf{69.88}          \\
                                  & iterative        & 55.24  &   89.89     & 71.72          & 69.68 \\
                                  & center-to-corner & \textbf{65.77}$\uparrow$ & \textbf{89.97}$\uparrow$ &  \textbf{71.73}$\uparrow$ & 69.41$\uparrow$ \\
      \bottomrule
    \end{tabular}
  \end{center}
\end{table*}

\noindent As aforementioned, the use of a quadrilateral representation for oriented
bounding boxes provides additional freedom in the way the regression head can
produce the final corners. We compare the performance of the different corner
prediction strategies we have introduced against the ``direct'' strategy, i.e.
the de facto standard approach in object detection.
\tablename~\ref{tab:exp:prediction-strategies} reports the experimental results
obtained on the validation sets for each prediction strategy with DAFNe (without
employing the oriented center-ness). When compared to the direct strategy, the
center-to-corner strategy improves performance on all datasets, by 0.14\%, 0.50\%,
1.76\%, and 4.09\% mAP for UCAS-AOD, DOTA 1.5, DOTA 1.0, and HRSC2016 respectively. The
offset and iterative prediction strategies lead to mixed results, turning the
center-to-corner strategy into the best choice since it outperforms the other
introduced strategies on UCAS-AOD, HRSC2016, and DOTA 1.0 and it is on par on DOTA 1.5.
This suggests that the division of predicting the object's center first and the
offset to its corners afterward does lead to better object localization. However,
the center-to-corner approach introduces a second branch, namely the center
regression branch, and thus, it requires 2.4M additional parameters. Therefore,
we investigate whether the increased model capacity is the
reason for this improvement. We evaluate the model on DOTA 1.0 using the direct
corner prediction strategy and modulate the tower's capacity by increasing the
number of convolution layers in the prediction branches from 1 to 8, resulting
in an increase of 1.2M parameters per layer.
\begin{figure}
  \centering \includegraphics{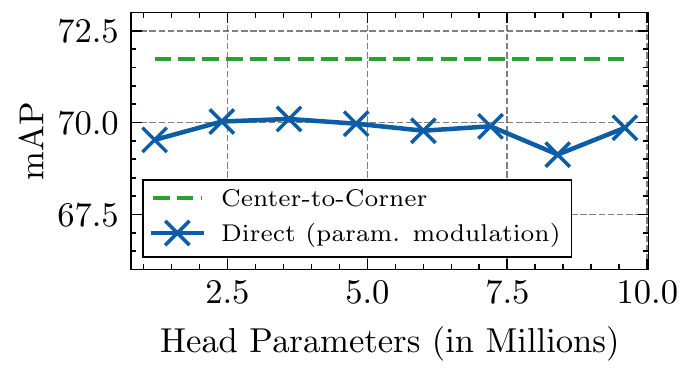}
  \caption{Modulation of the number of head convolutions for the classification
    and bounding box regression branches with the ``direct'' corner prediction
    strategy on the DOTA 1.0 validation set. Simply increasing the model
    capacity does not lead to the same improvements in accuracy as obtained with the
    introduced ``center-to-corner'' strategy.}
  \label{fig:exp:conv-params}
\end{figure}
As one can see in Figure~\ref{fig:exp:conv-params}, the best model achieved
70.10\% mAP (3 convolution layers), while the others performed slightly worse.
Thus, simply increasing the tower's capacity does not lead to better model
accuracy. The improvement observed with the center-to-corner regression approach
shows that it is a valid methodological advancement over the direct corner
regression. In summary, we can answer (\textbf{Q2}) positively, the introduced
divide-and-conquer corner prediction strategies outperform the direct approach.
The center-to-corner is, in general, the best choice.

\begin{table*}[h!]
  \caption{Test set accuracies (in mAP) of DAFNe compared with
    all other one-stage anchor-free oriented object detection models in the
    literature and with more complex two-stage and anchor-based models.
    Note that ``\mstrain'' indicates multi-scale
    training and testing, and ``\twostage'' indicates a two-stage
    model. We abbreviate backbone architectures with
    R(X)-50/101/152 as ResN(X)et-50/101/152, HG-104 as Hourglass-104, and DN-53
    as Darknet-53. Prefix ``Re'' indicates a rotation-equivariant
    architecture. Best results in bold, our approach is highlighted}
  \label{tab:exp:comparison}
  \scriptsize
  \begin{subtable}[t]{0.5\textwidth}
        \caption{Evaluation on DOTA 1.0}
    \label{tab:exp:comparison:dota10}
    \begin{center}
      \begin{tabular}{llc}
        \toprule
        Model                                & Backbone                                 & mAP                     \\
        \midrule
        \multicolumn{2}{l}{\textbf{Two-stage or                                         Anchor-based}}            &                                 \\
        APE\mstrain\twostage                 \cite{zhu2020ape}                          & RX-101                  & 75.75          \\
        OWSR\mstrain\twostage                \cite{Li_2019_CVPR_Workshops}              & R-101                   + RX-101         & 76.36 \\
                                             & +                                  mdcn-R-101                                 &                \\
        KLD                                  (RetinaNet)\mstrain                        \cite{Yang2021LearningHB} & R-50           & 78.32 \\
        ReDet\mstrain\twostage               \cite{han2021ReDet}                        & ReR-101+ReFPN                            & 80.10                   \\
        KLD                                  (R\textsuperscript{3}Det)\mstrain\twostage \cite{Yang2021LearningHB} & R-152                   & 80.17          \\
        Oriented                             R-CNN\mstrain\twostage                     \cite{Xie2021OrientedRF}  & R-50           & \textbf{80.87} \\
        \midrulethin
        \multicolumn{2}{l}{\textbf{One-stage and                                        Anchor-free}}             &                                 \\
        IENet\mstrain                        \cite{lin2019ienet}                        & \rnM                    & 57.15          \\
        PIoU\mstrain                         \cite{chen2020piou}                        & \dlaS                   & 60.50          \\
        Axis                                 Learning                                   \cite{xiao2020axis}       & \rnM           & 65.98 \\
        P-RSDet\mstrain                      \cite{zhou2020polar}                       & \rnM                    & 72.30          \\
        O\textsuperscript{2}-DNet            \cite{wei2020o2-dnet}                      & \hgM                    & 71.04          \\
        AF-EMS                               \cite{jia2021af-ems}                       & \rnM                    & 69.86          \\
        AROA                                 \cite{he2021aroa}                          & \rnM                    & 75.41          \\
        MEAD                                 \cite{He2021MEADAM}                        & \rnM                    & 74.80          \\
        CenterRot                            \cite{wang2021centerrot}                   & \rnM                    & 74.00          \\
        BBAV\mstrain                         \cite{Yi2021OrientedOD}                    & \rnM                    & 75.36          \\
        CFA\mstrain                          \cite{Guo_2021_CVPR}                       & \rnM                    & 75.05          \\
        \rowcolor{rowhighlightcolor} DAFNe\mstrain                                                             & \rnS                    & 76.73          \\
        \rowcolor{rowhighlightcolor} DAFNe\mstrain                                                             & \rnM                    & \textbf{76.95} \\
        \bottomrule
      \end{tabular}
    \end{center}

  \end{subtable}%
  \begin{subtable}[t]{0.5\textwidth}
        \caption{Evaluation on HRSC2016}
    \label{tab:exp:comparison:hrsc}
    \begin{center}
      \begin{tabular}{llc}
        \toprule
        Model                                & Backbone                                 & mAP                     \\
        \midrule
        \multicolumn{2}{l}{\textbf{Two-stage or                                         Anchor-based}}            &                                 \\
        KLD                                  (RetinaNet)\mstrain                        \cite{Yang2021LearningHB} & R-50           & 85.56 \\
        KLD                                  (R\textsuperscript{3}Det)\mstrain\twostage \cite{Yang2021LearningHB} & R-101                   & 87.45          \\
        ReDet\twostage                       \cite{han2021ReDet}                        & ReR-50+ReFPN            & 90.46          \\
        Oriented                             R-CNN\mstrain\twostage                     \cite{Xie2021OrientedRF}  & R-101          & \textbf{90.50} \\
        \midrulethin
        \multicolumn{2}{l}{\textbf{One-stage and                                        Anchor-free}}             &                                 \\
        IENet\mstrain                        \cite{lin2019ienet}                        & \rnM                    & 75.01          \\
        Axis                                 Learning                                   \cite{xiao2020axis}       & \rnM           & 78.51 \\
        PIoU\mstrain                        \cite{chen2020piou}                        & \rnM                    & 80.32          \\
        BBAV\mstrain                         \cite{Yi2021OrientedOD}                    & \rnM                    & 88.60          \\
        PIoU\mstrain                         \cite{chen2020piou}                        & \dlaS                   & 89.20          \\
        \rowcolor{rowhighlightcolor} DAFNe\mstrain                                                             & \rnM                    & 89.51          \\
        \rowcolor{rowhighlightcolor} DAFNe\mstrain                                                             & \rnS                    & 89.76          \\
        MEAD~\cite{He2021MEADAM}                                 & \rnM                                     & 89.83                   \\
        CenterRot                            \cite{wang2021centerrot}                   & \rnS                    & \textbf{90.20} \\
        \bottomrule
      \end{tabular}
    \end{center}

  \end{subtable}
  \begin{subtable}[t]{0.5\textwidth}
    
    \caption{Evaluation on DOTA 1.5}
    \label{tab:exp:comparison:dota15}
    \begin{center}
      \begin{tabular}{llc}
        \toprule
        Model                                & Backbone                                 & mAP                     \\
        \midrule
        \multicolumn{2}{l}{\textbf{Two-stage or                                         Anchor-based}}            &                                 \\
        OWSR\mstrain\twostage                \cite{Li_2019_CVPR_Workshops}              & R-101                   + RX-101         & 76.60 \\
                                             & +                                  mdcn-R-101                                 &                \\
        ReDet\mstrain\twostage \cite{han2021ReDet}               & ReR-50+ReFPN                             & 78.08                   \\
        APE\mstrain\twostage                 \cite{zhu2020ape}                          & RX-101                  & \textbf{78.34}          \\
        \midrulethin
        \multicolumn{2}{l}{\textbf{One-stage and                                        Anchor-free}}             &                                 \\
        \rowcolor{rowhighlightcolor} DAFNe\mstrain                                                             & \rnS                    & 71.47          \\
        \rowcolor{rowhighlightcolor} DAFNe\mstrain                                                             & \rnM                    & \textbf{71.99}          \\
        \bottomrule
      \end{tabular}
    \end{center}

  \end{subtable}%
  \begin{subtable}[t]{0.5\textwidth}
    
    \caption{Evaluation on UCAS-AOD}
    \label{tab:exp:comparison:ucas-aod}
    \begin{center}
      \begin{tabular}{llc}
        \toprule
        Model                                & Backbone                                 & mAP                     \\
        \midrule
        \multicolumn{2}{l}{\textbf{Two-stage or                                         Anchor-based}}            &                                 \\
        R-Yolov3                             \cite{yolo-v3}                             & \darknetS               & 82.08          \\
        Faster                               R-CNN\twostage                             \cite{faster-rcnn}        & \rnS           & 88.86 \\
        RoI                                  Transformer\twostage                       \cite{roi-trans}          & \rnS           & 89.02 \\
        SLA                                  \cite{ming2021sla}                         & \rnS                    & 89.44          \\
        CFC-Net                              \cite{ming2022cfcnet}                      & \rnS                    & 89.49          \\
        RIDet-O                              \cite{ridet}                               & \rnS                    & 89.62          \\
        DAL                                  \cite{Ming2021DynamicAL}                   & \rnS                    & 89.87          \\
        S2ANet                               \cite{han2020align}                        & \rnS                    & \textbf{89.99}          \\
        \midrulethin
        \multicolumn{2}{l}{\textbf{One-stage and                                        Anchor-free}}             &                                 \\
        R-RetinaNet                          \cite{retinanet}                           & \rnS                    & 87.57          \\
        \rowcolor{rowhighlightcolor} DAFNe\mstrain                                                             & \rnS                    & 89.73          \\
        \rowcolor{rowhighlightcolor} DAFNe\mstrain                                                             & \rnM                    & \textbf{89.85}          \\
        \bottomrule
      \end{tabular}
    \end{center}

  \end{subtable}
\end{table*}

\subsection{(Q3) DAFNe: The State-of-the-Art for One-Stage Anchor-Free Oriented
  Object Detection}
\label{sec:exp:comp-others}
One-stage anchor-free, as well as oriented object detection, are relatively new
in the field of computer vision. Only little research has been done at their
intersection. We compare DAFNe equipped with oriented center-ness,
center-to-corner prediction strategy, and the learning objectives as described
before with all other one-stage anchor-free methods in the literature, to the
best of our knowledge. Moreover, we compare it with the most prominent
two-stage and anchor-based approaches in the literature employed as gold standard.
Models are evaluated on the test sets of UCAS-AOD,
HRSC2016, DOTA 1.0, DOTA 1.5. In \tablename~\ref{tab:exp:comparison:dota10} we
show that DAFNe outperforms all previous one-stage anchor-free models on DOTA
1.0 by a large margin, setting the new state-of-the-art result of 76.95\% mAP.
For a comprehensive class-separated evaluation on DOTA 1.0, see Appendix C.
On HRSC2016 (\tablename~\ref{tab:exp:comparison:hrsc}), DAFNe performs close to the current best model, CenterRot~\cite{wang2021centerrot}. Since no
other one-stage anchor-free model has been evaluated on DOTA 1.5 (\tablename~\ref{tab:exp:comparison:dota15}), we set a new
baseline of 71.99\% mAP. We outperform R-RetinaNet on UCAS-AOD (\tablename~\ref{tab:exp:comparison:ucas-aod}) by 2.28\% mAP and perform only
0.14\% mAP below the state-of-the-art result for \textit{anchor-based} models (S2ANet,
89.99\% mAP). In general, being more accurate, DAFNe reduces the gap with more
complex two-stage and anchor-based models. Therefore, we can answer (\textbf{Q3})
affirmatively, DAFNe sets the new state-of-the-art accuracy for one-stage anchor-free
oriented object detection.

\section{Conclusion}
\label{sec:conclusion}
In this work, we have introduced DAFNe, a one-stage anchor-free approach for
oriented object detection, a central task in computer vision. Taking into
account the orientation of both the objects and the bounding boxes makes
oriented object detection a more general and, thus, a harder task than the
simpler horizontal one.
Compared to two-stage anchor-based solutions, DAFNe is simpler in design,
faster, and easier to optimize, and it does not need data-specific anchors. To
overcome the additional challenges of oriented object detection, next to
presenting DAFNe's design and learning objectives that foster training stability,
we have introduced a novel oriented center-ness and new divide-and-conquer
corner prediction strategies. We have demonstrated how our oriented center-ness
substantially improves accuracy. Moreover, we have shown that more advanced
corner prediction strategies can improve localization performance, 
with the proposed ``center-to-corner'' strategy consistently outperforming the ``direct'' baseline strategy.
Potentially, these can be beneficial also for the horizontal case. With these
contributions, we have set the new state-of-the-art accuracy for one-stage
anchor-free oriented object detection on the DOTA 1.0, DOTA 1.5, and
UCAS-AOD benchmarks and perform on par with the previous best models on HRSC2016.
Since recent anchor-based models are currently the gold standard in terms of
accuracy, DAFNe presents a step forward to close the gap with these more complex
solutions.

\section* {Acknowledgements}
This work was supported by the Federal Ministry of Education and Research
(BMBF) Competence Center for AI and Labour ``kompAKI'' (FKZ 02L19C150), the
German Science Foundation (DFG, German Research Foundation; GRK 1994/1
``AIPHES''), and the Hessian Ministry of Higher Education, Research, Science
and the Arts (HMWK) project ``The Third Wave of AI''.

\bibliography{./aaai22.bib}

\begin{thebibliography}{60}
\providecommand{\natexlab}[1]{#1}

\bibitem[{Chen et~al.(2017)Chen, Papandreou, Schroff, and
  Adam}]{Chen2017RethinkingAC}
Chen, L.-C.; Papandreou, G.; Schroff, F.; and Adam, H. 2017.
\newblock Rethinking Atrous Convolution for Semantic Image Segmentation.
\newblock arXiv preprint arXiv: 1706.05587.
\newblock arXiv:1706.05587.

\bibitem[{Chen et~al.(2020)Chen, Chen, Lin, See, Yu, Ke, and
  Yang}]{chen2020piou}
Chen, Z.; Chen, K.-A.; Lin, W.; See, J.; Yu, H.; Ke, Y.; and Yang, C. 2020.
\newblock PIoU Loss: Towards Accurate Oriented Object Detection in Complex
  Environments.
\newblock In \emph{European Conference on Computer Vision (ECCV)}.

\bibitem[{Dai, He, and Sun(2016)}]{instance-aware-semantiv-seg}
Dai, J.; He, K.; and Sun, J. 2016.
\newblock Instance-Aware Semantic Segmentation via Multi-task Network Cascades.
\newblock \emph{Conference on Computer Vision and Pattern Recognition (CVPR)}.

\bibitem[{Deng et~al.(2009)Deng, Dong, Socher, Li, Li, and
  Fei-Fei}]{imagenet_cvpr09}
Deng, J.; Dong, W.; Socher, R.; Li, L.-J.; Li, K.; and Fei-Fei, L. 2009.
\newblock {ImageNet: A Large-Scale Hierarchical Image Database}.
\newblock In \emph{Conference on Computer Vision and Pattern Recognition
  (CVPR)}.

\bibitem[{Ding et~al.(2019)Ding, Xue, Long, Xia, and Lu}]{roi-trans}
Ding, J.; Xue, N.; Long, Y.; Xia, G.-S.; and Lu, Q. 2019.
\newblock Learning RoI Transformer for Oriented Object Detection in Aerial
  Images.
\newblock In \emph{Conference on Computer Vision and Pattern Recognition
  (CVPR)}.

\bibitem[{Duan et~al.(2019)Duan, Bai, Xie, Qi, Huang, and Tian}]{centernet}
Duan, K.; Bai, S.; Xie, L.; Qi, H.; Huang, Q.; and Tian, Q. 2019.
\newblock CenterNet: Keypoint Triplets for Object Detection.
\newblock In \emph{International Conference on Computer Vision (ICCV)}.

\bibitem[{Goodfellow, Bengio, and Courville(2016)}]{goodfellow2016deeplearning}
Goodfellow, I.; Bengio, Y.; and Courville, A. 2016.
\newblock \emph{Deep Learning}.
\newblock The MIT Press.

\bibitem[{Goyal et~al.(2018)Goyal, Dollár, Girshick, Noordhuis, Wesolowski,
  Kyrola, Tulloch, Jia, and He}]{goyal2017accurateLM}
Goyal, P.; Dollár, P.; Girshick, R.; Noordhuis, P.; Wesolowski, L.; Kyrola,
  A.; Tulloch, A.; Jia, Y.; and He, K. 2018.
\newblock Accurate, Large Minibatch SGD: Training ImageNet in 1 Hour.
\newblock arXiv preprint arXiv: 1706.02677.
\newblock arXiv:1706.02677.

\bibitem[{Guo et~al.(2021)Guo, Liu, Zhang, Jiao, Ji, and Ye}]{Guo_2021_CVPR}
Guo, Z.; Liu, C.; Zhang, X.; Jiao, J.; Ji, X.; and Ye, Q. 2021.
\newblock Beyond Bounding-Box: Convex-Hull Feature Adaptation for Oriented and
  Densely Packed Object Detection.
\newblock In \emph{Conference on Computer Vision and Pattern Recognition
  (CVPR)}.

\bibitem[{Han et~al.(2021{\natexlab{a}})Han, Ding, Li, and Xia}]{han2020align}
Han, J.; Ding, J.; Li, J.; and Xia, G.-S. 2021{\natexlab{a}}.
\newblock Align Deep Features for Oriented Object Detection.
\newblock \emph{IEEE Trans. on Geoscience and Remote Sensing}.

\bibitem[{Han et~al.(2021{\natexlab{b}})Han, Ding, Xue, and Xia}]{han2021ReDet}
Han, J.; Ding, J.; Xue, N.; and Xia, G.-S. 2021{\natexlab{b}}.
\newblock ReDet: A Rotation-equivariant Detector for Aerial Object Detection.
\newblock In \emph{Proceedings of the IEEE/CVF Conference on Computer Vision
  and Pattern Recognition (CVPR)}.

\bibitem[{Hariharan et~al.(2014)Hariharan, Arbel{\'a}ez, Girshick, and
  Malik}]{simul-det-and-seg}
Hariharan, B.; Arbel{\'a}ez, P.; Girshick, R.; and Malik, J. 2014.
\newblock Simultaneous Detection and Segmentation.
\newblock In \emph{European Conference on Computer Vision (ECCV)}.

\bibitem[{Hariharan et~al.(2015)Hariharan, Arbeláez, Girshick, and
  Malik}]{hypercolumns}
Hariharan, B.; Arbeláez, P.; Girshick, R.; and Malik, J. 2015.
\newblock Hypercolumns for object segmentation and fine-grained localization.
\newblock In \emph{Conference on Computer Vision and Pattern Recognition
  (CVPR)}.

\bibitem[{He et~al.(2017)He, Gkioxari, Dollár, and Girshick}]{mask-rcnn}
He, K.; Gkioxari, G.; Dollár, P.; and Girshick, R. 2017.
\newblock Mask R-CNN.
\newblock In \emph{International Conference on Computer Vision (ICCV)}.

\bibitem[{He et~al.(2016)He, Zhang, Ren, and Sun}]{resnet}
He, K.; Zhang, X.; Ren, S.; and Sun, J. 2016.
\newblock Deep Residual Learning for Image Recognition.
\newblock In \emph{Conference on Computer Vision and Pattern Recognition
  (CVPR)}.

\bibitem[{He et~al.(2021{\natexlab{a}})He, Ma, He, Zhang, Liu, and
  Ru}]{he2021aroa}
He, X.; Ma, S.; He, L.; Zhang, F.; Liu, X.; and Ru, L. 2021{\natexlab{a}}.
\newblock AROA: Attention Refinement One-Stage Anchor-Free Detector for Objects
  in Remote Sensing Imagery.
\newblock In \emph{Image and Graphics}. Springer International Publishing.

\bibitem[{He et~al.(2021{\natexlab{b}})He, Ren, Yang, Yang, and
  Zhang}]{He2021MEADAM}
He, Z.; Ren, Z.; Yang, X.; Yang, Y.; and Zhang, W. 2021{\natexlab{b}}.
\newblock MEAD: a Mask-guidEd Anchor-free Detector for oriented aerial object
  detection.
\newblock \emph{Applied Intelligence}.

\bibitem[{Kang et~al.(2018)Kang, Li, Yan, Zeng, Yang, Xiao, Zhang, Wang, Wang,
  Wang, and Ouyang}]{8003302}
Kang, K.; Li, H.; Yan, J.; Zeng, X.; Yang, B.; Xiao, T.; Zhang, C.; Wang, Z.;
  Wang, R.; Wang, X.; and Ouyang, W. 2018.
\newblock T-CNN: Tubelets With Convolutional Neural Networks for Object
  Detection From Videos.
\newblock \emph{IEEE Trans. on Circuits and Systems for Video Technology}, 28.

\bibitem[{Karpathy and Fei-Fei(2015)}]{karpathy2015deep}
Karpathy, A.; and Fei-Fei, L. 2015.
\newblock Deep visual-semantic alignments for generating image descriptions.
\newblock In \emph{Conference on Computer Vision and Pattern Recognition
  (CVPR)}.

\bibitem[{Kong et~al.(2019)Kong, Sun, Liu, Jiang, and Shi}]{kong2019foveabox}
Kong, T.; Sun, F.; Liu, H.; Jiang, Y.; and Shi, J. 2019.
\newblock FoveaBox: Beyond Anchor-based Object Detector.
\newblock arXiv preprint arXiv: 1904.03797.
\newblock arXiv:1904.03797.

\bibitem[{Law and Deng(2019)}]{cornernet}
Law, H.; and Deng, J. 2019.
\newblock CornerNet: Detecting Objects as Paired Keypoints.
\newblock \emph{International Journal of Computer Vision}.

\bibitem[{Li et~al.(2019{\natexlab{a}})Li, Xu, Cui, Wang, Jie, Zhang, and
  Yang}]{Li_2019_CVPR_Workshops}
Li, C.; Xu, C.; Cui, Z.; Wang, D.; Jie, Z.; Zhang, T.; and Yang, J.
  2019{\natexlab{a}}.
\newblock Learning Object-Wise Semantic Representation for Detection in Remote
  Sensing Imagery.
\newblock In \emph{CVPR Workshops}.

\bibitem[{Li et~al.(2019{\natexlab{b}})Li, Xu, Cui, Wang, Zhang, and
  Yang}]{li2019ucas-aod}
Li, C.; Xu, C.; Cui, Z.; Wang, D.; Zhang, T.; and Yang, J. 2019{\natexlab{b}}.
\newblock Feature-Attentioned Object Detection in Remote Sensing Imagery.
\newblock In \emph{2019 IEEE International Conference on Image Processing
  (ICIP)}.

\bibitem[{Lin et~al.(2017)Lin, Dollár, Girshick, He, Hariharan, and
  Belongie}]{fpns}
Lin, T.-Y.; Dollár, P.; Girshick, R.; He, K.; Hariharan, B.; and Belongie, S.
  2017.
\newblock Feature Pyramid Networks for Object Detection.
\newblock In \emph{Conference on Computer Vision and Pattern Recognition
  (CVPR)}.

\bibitem[{Lin et~al.(2020)Lin, Goyal, Girshick, He, and Dollár}]{retinanet}
Lin, T.-Y.; Goyal, P.; Girshick, R.; He, K.; and Dollár, P. 2020.
\newblock Focal Loss for Dense Object Detection.
\newblock \emph{IEEE Trans. on Pattern Analysis and Machine Intelligence}, 42.

\bibitem[{Lin, Feng, and Guan(2019)}]{lin2019ienet}
Lin, Y.; Feng, P.; and Guan, J. 2019.
\newblock IENet: Interacting Embranchment One Stage Anchor Free Detector for
  Orientation Aerial Object Detection.
\newblock arXiv preprint arXiv: 1912.00969.
\newblock arXiv:1912.00969.

\bibitem[{Liu, Pan, and Lei(2017)}]{Liu2017LearningAR}
Liu, L.; Pan, Z.; and Lei, B. 2017.
\newblock Learning a Rotation Invariant Detector with Rotatable Bounding Box.
\newblock arXiv preprint arXiv: 1711.09405.
\newblock arXiv:1711.09405.

\bibitem[{Liu et~al.(2016)Liu, Anguelov, Erhan, Szegedy, Reed, Fu, and
  Berg}]{ssd}
Liu, W.; Anguelov, D.; Erhan, D.; Szegedy, C.; Reed, S.~E.; Fu, C.-Y.; and
  Berg, A.~C. 2016.
\newblock SSD: Single Shot MultiBox Detector.
\newblock In \emph{European Conference on Computer Vision (ECCV)}.

\bibitem[{Liu et~al.(2017)Liu, Yuan, Weng, and Yang}]{hrsc2016}
Liu, Z.; Yuan, L.; Weng, L.; and Yang, Y. 2017.
\newblock A High Resolution Optical Satellite Image Dataset for Ship
  Recognition and Some New Baselines.

\bibitem[{Ma et~al.(2018)Ma, Shao, Ye, Wang, Wang, Zheng, and
  Xue}]{Jianqi17RRPN}
Ma, J.; Shao, W.; Ye, H.; Wang, L.; Wang, H.; Zheng, Y.; and Xue, X. 2018.
\newblock Arbitrary-Oriented Scene Text Detection via Rotation Proposals.
\newblock \emph{IEEE Trans. on Multimedia}, 20.

\bibitem[{Ming et~al.(2022{\natexlab{a}})Ming, Miao, Zhou, and
  Dong}]{ming2022cfcnet}
Ming, Q.; Miao, L.; Zhou, Z.; and Dong, Y. 2022{\natexlab{a}}.
\newblock CFC-Net: A Critical Feature Capturing Network for Arbitrary-Oriented
  Object Detection in Remote-Sensing Images.
\newblock \emph{IEEE Transactions on Geoscience and Remote Sensing}, 60.

\bibitem[{Ming et~al.(2021{\natexlab{a}})Ming, Miao, Zhou, Song, and
  Yang}]{ming2021sla}
Ming, Q.; Miao, L.; Zhou, Z.; Song, J.; and Yang, X. 2021{\natexlab{a}}.
\newblock Sparse Label Assignment for Oriented Object Detection in Aerial
  Images.
\newblock \emph{Remote Sensing}, 13.

\bibitem[{Ming et~al.(2022{\natexlab{b}})Ming, Miao, Zhou, Yang, and
  Dong}]{ridet}
Ming, Q.; Miao, L.; Zhou, Z.; Yang, X.; and Dong, Y. 2022{\natexlab{b}}.
\newblock Optimization for Arbitrary-Oriented Object Detection via
  Representation Invariance Loss.
\newblock \emph{IEEE Geoscience and Remote Sensing Letters}, 19.

\bibitem[{Ming et~al.(2021{\natexlab{b}})Ming, Zhou, Miao, Zhang, and
  Li}]{Ming2021DynamicAL}
Ming, Q.; Zhou, Z.; Miao, L.; Zhang, H.; and Li, L. 2021{\natexlab{b}}.
\newblock Dynamic Anchor Learning for Arbitrary-Oriented Object Detection.
\newblock In \emph{Association for the Advancement of Artificial Intelligence
  (AAAI)}.

\bibitem[{Qian et~al.(2021)Qian, Yang, Peng, Guo, and Yan}]{qian2019learning}
Qian, W.; Yang, X.; Peng, S.; Guo, Y.; and Yan, J. 2021.
\newblock Learning Modulated Loss for Rotated Object Detection.
\newblock In \emph{Association for the Advancement of Artificial Intelligence
  (AAAI)}.

\bibitem[{Redmon et~al.(2016)Redmon, Divvala, Girshick, and Farhadi}]{yolo}
Redmon, J.; Divvala, S.; Girshick, R.; and Farhadi, A. 2016.
\newblock You Only Look Once: Unified, Real-Time Object Detection.
\newblock In \emph{Conference on Computer Vision and Pattern Recognition
  (CVPR)}.

\bibitem[{Redmon and Farhadi(2017)}]{yolo-v2}
Redmon, J.; and Farhadi, A. 2017.
\newblock YOLO9000: Better, Faster, Stronger.
\newblock In \emph{Conference on Computer Vision and Pattern Recognition
  (CVPR)}.

\bibitem[{Redmon and Farhadi(2018)}]{yolo-v3}
Redmon, J.; and Farhadi, A. 2018.
\newblock YOLOv3: An Incremental Improvement.
\newblock arXiv preprint arXiv: 1804.02767.
\newblock arXiv:1804.02767.

\bibitem[{Ren et~al.(2015)Ren, He, Girshick, and Sun}]{faster-rcnn}
Ren, S.; He, K.; Girshick, R.; and Sun, J. 2015.
\newblock Faster R-CNN: Towards Real-Time Object Detection with Region Proposal
  Networks.
\newblock In \emph{Advances in Neural Information Processing Systems
  ({NeurIPS})}.

\bibitem[{Sun et~al.(2018)Sun, Chen, Luke, and Shang}]{sun2018salience}
Sun, P.; Chen, G.; Luke, G.; and Shang, Y. 2018.
\newblock Salience Biased Loss for Object Detection in Aerial Images.
\newblock arXiv preprint arXiv: 1810.08103.
\newblock arXiv:1810.08103.

\bibitem[{Tian et~al.(2019)Tian, Shen, Chen, and He}]{fcos}
Tian, Z.; Shen, C.; Chen, H.; and He, T. 2019.
\newblock {FCOS}: Fully Convolutional One-Stage Object Detection.
\newblock In \emph{International Conference on Computer Vision (ICCV)}.

\bibitem[{Wang, Yang, and Li(2021)}]{wang2021centerrot}
Wang, J.; Yang, L.; and Li, F. 2021.
\newblock Predicting Arbitrary-Oriented Objects as Points in Remote Sensing
  Images.
\newblock \emph{Remote Sensing}, 13.

\bibitem[{Wei et~al.(2020)Wei, Zhang, Chang, Li, Wang, and
  Sun}]{wei2020o2-dnet}
Wei, H.; Zhang, Y.; Chang, Z.; Li, H.; Wang, H.; and Sun, X. 2020.
\newblock Oriented objects as pairs of Middle Lines.
\newblock \emph{ISPRS Journal of Photogrammetry and Remote Sensing}, 169.

\bibitem[{Wu et~al.(2018)Wu, Shen, Wang, Dick, and van~den
  Hengel}]{Wu2018ImageCA}
Wu, Q.; Shen, C.; Wang, P.; Dick, A.; and van~den Hengel, A. 2018.
\newblock Image Captioning and Visual Question Answering Based on Attributes
  and External Knowledge.
\newblock \emph{IEEE Trans. on Pattern Analysis and Machine Intelligence}, 40.

\bibitem[{Xia et~al.(2018)Xia, Bai, Ding, Zhu, Belongie, Luo, Datcu, Pelillo,
  and Zhang}]{Xia_2018_CVPR}
Xia, G.-S.; Bai, X.; Ding, J.; Zhu, Z.; Belongie, S.; Luo, J.; Datcu, M.;
  Pelillo, M.; and Zhang, L. 2018.
\newblock DOTA: A Large-Scale Dataset for Object Detection in Aerial Images.
\newblock In \emph{Conference on Computer Vision and Pattern Recognition
  (CVPR)}.

\bibitem[{Xiao et~al.(2020)Xiao, Qian, Shao, Tan, and Wang}]{xiao2020axis}
Xiao, Z.; Qian, L.; Shao, W.; Tan, X.; and Wang, K. 2020.
\newblock Axis Learning for Orientated Objects Detection in Aerial Images.
\newblock \emph{Remote Sensing}, 12.

\bibitem[{Xie et~al.(2021)Xie, Cheng, Wang, Yao, and Han}]{Xie2021OrientedRF}
Xie, X.; Cheng, G.; Wang, J.; Yao, X.; and Han, J. 2021.
\newblock Oriented R-CNN for Object Detection.
\newblock \emph{International Conference on Computer Vision (ICCV)}.

\bibitem[{Xu et~al.(2015)Xu, Ba, Kiros, Cho, Courville, Salakhudinov, Zemel,
  and Bengio}]{pmlr-v37-xuc15}
Xu, K.; Ba, J.; Kiros, R.; Cho, K.; Courville, A.; Salakhudinov, R.; Zemel, R.;
  and Bengio, Y. 2015.
\newblock Show, Attend and Tell: Neural Image Caption Generation with Visual
  Attention.
\newblock In \emph{International Conference on Machine Learning (ICML)},
  volume~37.

\bibitem[{Yan et~al.(2021)Yan, Zhao, Diao, Wang, and Sun}]{jia2021af-ems}
Yan, J.; Zhao, L.; Diao, W.; Wang, H.; and Sun, X. 2021.
\newblock AF-EMS Detector: Improve the Multi-Scale Detection Performance of the
  Anchor-Free Detector.
\newblock \emph{Remote Sensing}, 13.

\bibitem[{Yang et~al.(2021{\natexlab{a}})Yang, Liu, Yan, and
  Li}]{Yang2021R3DetRS}
Yang, X.; Liu, Q.; Yan, J.; and Li, A. 2021{\natexlab{a}}.
\newblock R3Det: Refined Single-Stage Detector with Feature Refinement for
  Rotating Object.
\newblock In \emph{Association for the Advancement of Artificial Intelligence
  (AAAI)}.

\bibitem[{Yang and Yan(2020)}]{yang2020arbitrary}
Yang, X.; and Yan, J. 2020.
\newblock Arbitrary-Oriented Object Detection with Circular Smooth Label.
\newblock \emph{European Conference on Computer Vision (ECCV)}.

\bibitem[{Yang et~al.(2020)Yang, Yan, Yang, Tang, Liao, and
  He}]{yang2020scrdet}
Yang, X.; Yan, J.; Yang, X.; Tang, J.; Liao, W.; and He, T. 2020.
\newblock SCRDet++: Detecting Small, Cluttered and Rotated Objects via
  Instance-Level Feature Denoising and Rotation Loss Smoothing.
\newblock arXiv preprint arXiv: 2004.13316.
\newblock arXiv:2004.13316.

\bibitem[{Yang et~al.(2021{\natexlab{b}})Yang, Yang, Yang, Ming, Wang, Tian,
  and Yan}]{Yang2021LearningHB}
Yang, X.; Yang, X.; Yang, J.; Ming, Q.; Wang, W.; Tian, Q.; and Yan, J.
  2021{\natexlab{b}}.
\newblock Learning High-Precision Bounding Box for Rotated Object Detection via
  Kullback-Leibler Divergence.
\newblock \emph{Advances in Neural Information Processing Systems ({NeurIPS})}.

\bibitem[{Yi et~al.(2021)Yi, Wu, Liu, Huang, Qu, and
  Metaxas}]{Yi2021OrientedOD}
Yi, J.; Wu, P.; Liu, B.; Huang, Q.; Qu, H.; and Metaxas, D.~N. 2021.
\newblock Oriented Object Detection in Aerial Images with Box Boundary-Aware
  Vectors.
\newblock \emph{Winter Conference on Applications of Computer Vision (WACV)}.

\bibitem[{Yu et~al.(2016)Yu, Jiang, Wang, Cao, and Huang}]{unitbox}
Yu, J.; Jiang, Y.; Wang, Z.; Cao, Z.; and Huang, T. 2016.
\newblock UnitBox: An Advanced Object Detection Network.
\newblock In \emph{Proceedings of ACM International Conference on Multimedia}.

\bibitem[{Zhou et~al.(2020{\natexlab{a}})Zhou, Wei, Li, Zhao, Zhang, and
  Zhang}]{zhou2020polar}
Zhou, L.; Wei, H.; Li, H.; Zhao, W.; Zhang, Y.; and Zhang, Y.
  2020{\natexlab{a}}.
\newblock Arbitrary-Oriented Object Detection in Remote Sensing Images Based on
  Polar Coordinates.
\newblock \emph{IEEE Access}.

\bibitem[{Zhou et~al.(2020{\natexlab{b}})Zhou, Wei, Li, Zhao, Zhang, and
  Zhang}]{zhou2020objects}
Zhou, L.; Wei, H.; Li, H.; Zhao, W.; Zhang, Y.; and Zhang, Y.
  2020{\natexlab{b}}.
\newblock Objects detection for remote sensing images based on polar
  coordinates.
\newblock arXiv preprint arXiv: 2001.02988.
\newblock arXiv:2001.02988.

\bibitem[{Zhou, Zhuo, and Kr{\"a}henb{\"u}hl(2019)}]{extremenet}
Zhou, X.; Zhuo, J.; and Kr{\"a}henb{\"u}hl, P. 2019.
\newblock Bottom-up Object Detection by Grouping Extreme and Center Points.
\newblock In \emph{Conference on Computer Vision and Pattern Recognition
  (CVPR)}.

\bibitem[{Zhou et~al.(2017)Zhou, Ye, Qiu, and Jiao}]{orn}
Zhou, Y.; Ye, Q.; Qiu, Q.; and Jiao, J. 2017.
\newblock Oriented Response Networks.
\newblock In \emph{Conference on Computer Vision and Pattern Recognition
  (CVPR)}.

\bibitem[{Zhu, Du, and Wu(2020)}]{zhu2020ape}
Zhu, Y.; Du, J.; and Wu, X. 2020.
\newblock Adaptive Period Embedding for Representing Oriented Objects in Aerial
  Images.
\newblock \emph{IEEE Transactions on Geoscience and Remote Sensing}, 58.

\end{thebibliography}


\appendix
\section*{Appendix}

\section{DAFNe Architecture}

\begin{figure*}[!b]
  \centering \includegraphics[width=\linewidth]{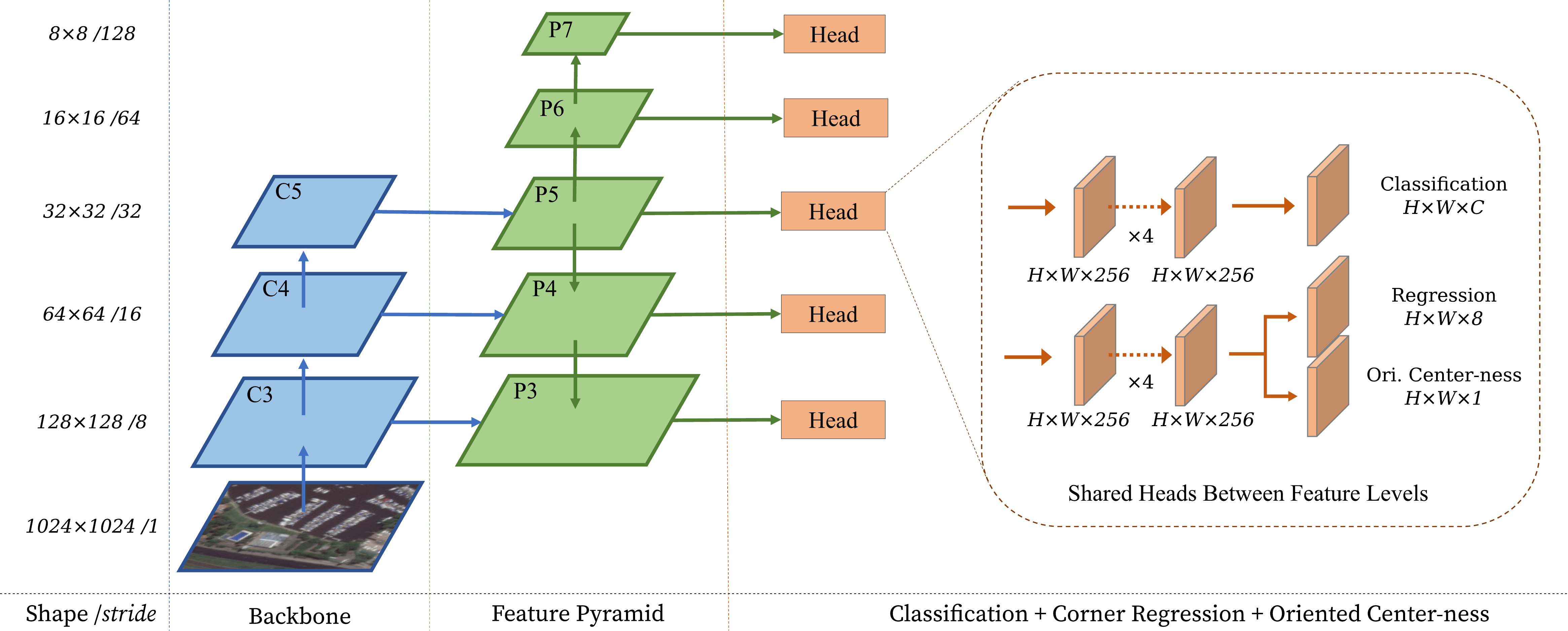}
  \caption{The main network architecture of DAFNe. $H \times W$ denotes
    the height and width of the feature maps and $/s$ is the down-sampling
    ratio w.r.t. the input size, also called \textit{stride}. The model uses
    feature maps \textit{C3}, \textit{C4}, and \textit{C5} of the backbone
    network, feeds them into levels \textit{P3-5} of the FPN and uses
    \textit{P3-7} for the final prediction of class confidences, object
    location, and oriented center-ness. The depiction is a modification of the one presented in \cite{fcos}. }
  \label{fig:method:base-architecture}
\end{figure*}

DAFNe is a one-stage anchor-free model for oriented object detection.
It is faster and easier to optimize than two-stage solutions and it
does not require predefined anchors as its anchor-based counterparts.
To date, it is the most accurate one-stage anchor-free model on DOTA
1.0, DOTA 1.5, and UCAS-AOD.
DAFNe architecture can be divided into three parts: 1) The backbone network, ResNet,
which is pre-trained to perform image classification on a large dataset such as ImageNet,
thus, it is able to extract meaningful and robust intermediate deep feature maps,
2) the Feature Pyramid Network which can construct
feature pyramids from the backbone feature maps using lateral top-down and
bottom-up connections with marginal cost and 3) the task-specific heads which
are separate sequential convolution branches on top of the FPN output that
produce predictions for different task objectives. A comprehensive illustration
of the architecture can be found in
\figurename~\ref{fig:method:base-architecture}.

In particular, the FPN extracts three feature maps, namely $C_{3}$,
$C_{4}$, and $C_{5}$ from the ResNet backbone.
As outlined in \figurename~\ref{fig:method:base-architecture}, $P_{3}$, $P_{4}$,
and $P_{5}$ are constructed using $C_{3}$, $C_{4}$, and $C_{5}$ while connecting
$P_{5}$ with $P_{4}$ as well as $P_{4}$ with $P_{3}$ using lateral connections
in a top-down fashion using nearest-neighbor upsampling and $1 \times 1$
convolutions. Pyramid level $P_{6}$ is obtained using a $3 \times 3$ convolution
with stride 2 on $P_{5}$, while level $P_{7}$ is obtained using a $3 \times 3$
convolution with stride 2 and a ReLU activation on $P_{6}$. Each pyramid level
has an additional $3 \times 3$ output convolution with 256 channels. This
results in five FPN output feature maps $\{P_{3}, P_{4}, P_{5}, P_{6}, P_{7}\}$
with dimensions $\frac{H}{2^{l}} \times \frac{W}{2^{l}} \times 256$ for output
$P_{l}$ at level $l$. The different feature maps
serve as deep representations of the input at different spatial resolutions.
Each feature in $P_{l}$ is responsible for patches of
\begin{align*}
  \frac{H}{\frac{H}{2^{l}}} \times \frac{W}{\frac{W}{2^{l}}} = 2^{l} \times 2^{l}
\end{align*}
pixels in the network input. In the context of object detection, this means that
each feature map $P_{l}$ is responsible for a different object size.
\begin{figure*}[!t]
  \centering \includegraphics[width=0.45\textwidth]{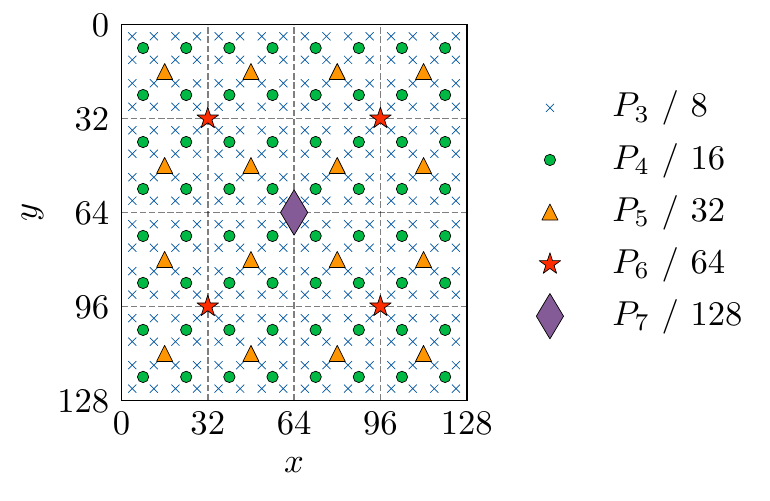}
  \caption{Feature map responsibilities w.r.t. an exemplary $128 \times 128 $
    pixel image input for FPN outputs $\{P_{3}, P_{4}, P_{5}, P_{6}, P_{7}\}$
    with their respective strides. For example, $P_{5}$
    has an output stride $s=32$, thus, a
    single feature in $P_{5}$ covers a block of $32 \times 32$ pixels in the
    input image, resulting in $\frac{128}{32} \cdot \frac{128}{32} = 16$
    features per channel. A feature map location $(x, y)$ can be mapped back
    onto the input image as follows:
    $(\lceil \frac{s}{2} \rceil + xs, \lceil \frac{s}{2} \rceil + ys)$.}
  \label{fig:method:locations}
\end{figure*}
\figurename~\ref{fig:method:locations} shows the mapping of feature map
responsibilities for each FPN output level onto an exemplary blank image as
input image of size $128 \times 128$.

Finally, the last architectural section, also called the \textit{head}, takes
care of the task-specific objectives. For this, each objective is being
assigned to a single branch in the head. A branch starts with a convolution
tower, i.e. a sequence of multiple $3 \times 3$ convolution layers that retain
the spatial dimensions of their input by padding the sides accordingly. On top
of the convolution tower, a task-specific $3 \times 3$ prediction
convolution is placed to map the tower outputs to the task-specific values. In this
work, we employ two branches as depicted in the rightmost section of
\figurename~\ref{fig:method:base-architecture}.

The first branch generates a $H \times W \times |C|$ class
prediction output, where $|C|$ corresponds to the number of classes present in
the dataset. In other words, for each location in the spatial grid of size
$H \times W$, the network generates a logits vector of length $|C|$ that
indicates the class of a possible object at that location, taking into
account also that there might not be an object. In this case, this
outcome is filtered out by employing a simple class score threshold later on.

The second branch produces a $H \times W \times 8$ coordinate regression output,
where each vector of length 8 in the $H \times W$ output map corresponds to an
oriented bounding box represented as quadrilateral of its four corner points
$(x_{0}, y_{0}, x_{1}, y_{1}, x_{2}, y_{2}, x_{3}, y_{3})$. Each coordinate is
interpreted as an offset vector relative to the location at which the
prediction occurs. Furthermore, a second $H \times W \times 1$ oriented center-ness
prediction is realized, which is then transformed into normalized oriented center-ness values
using the sigmoid transformation.

\section{Class-based Oriented Center-Ness Heatmaps}

\begin{figure*}[!h]
  \centering \includegraphics[width=\linewidth]{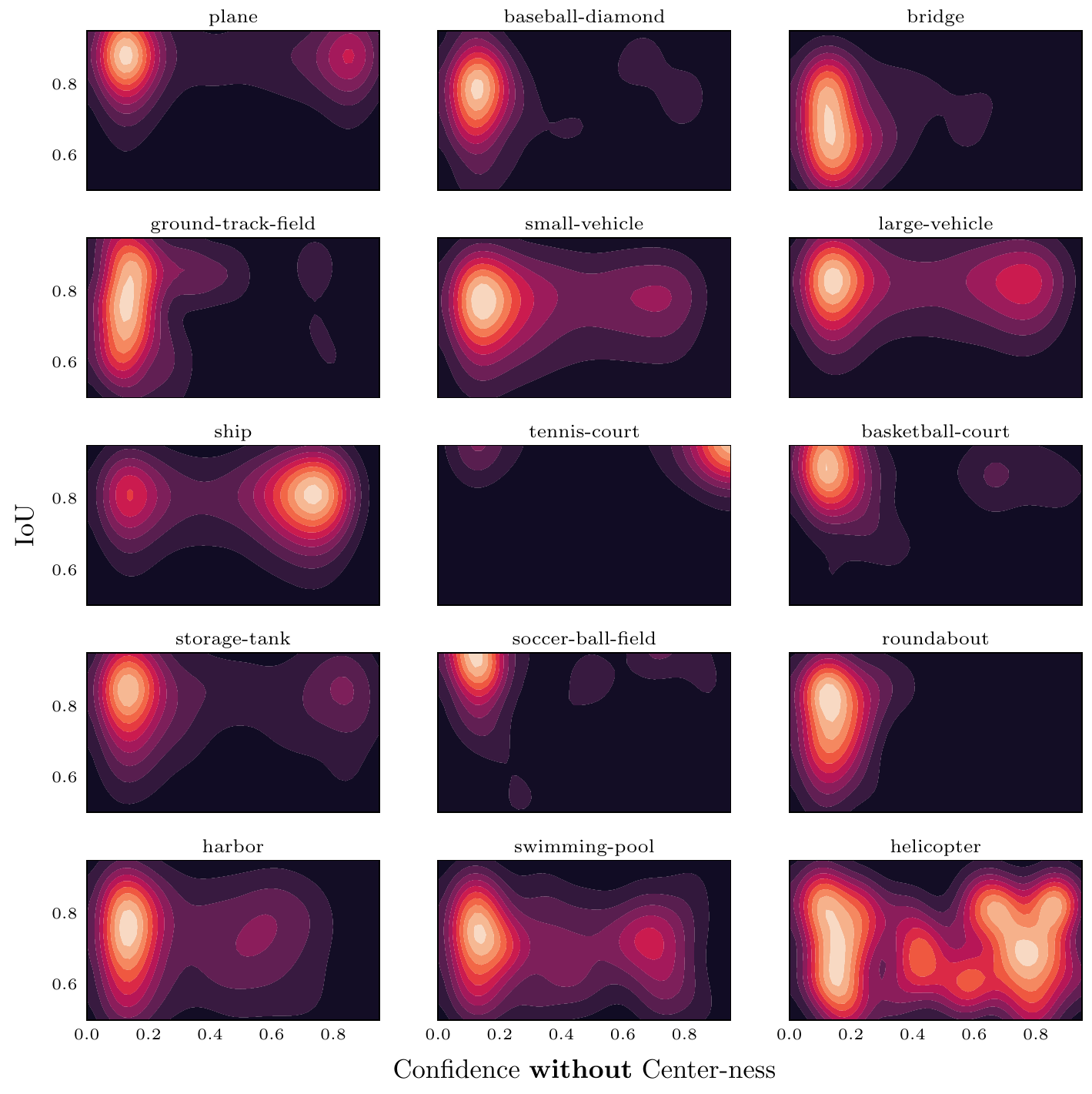}
  \caption{Heatmaps of classification confidences against IoU values for
    correctly detected oriented bounding boxes in the DOTA 1.0 validation set
    (true positives). Values are obtained from a model without oriented center-ness. In most
    of the classes the confidence concentrates around 0.1 and 0.2. This increases the chance that correct
    detections are wrongly discarded in the post-processing steps.}
  \label{fig:exp:oriented-center-ness-classes-baseline}
\end{figure*}

\begin{figure*}[!h]
  \centering
  \includegraphics[width=\linewidth]{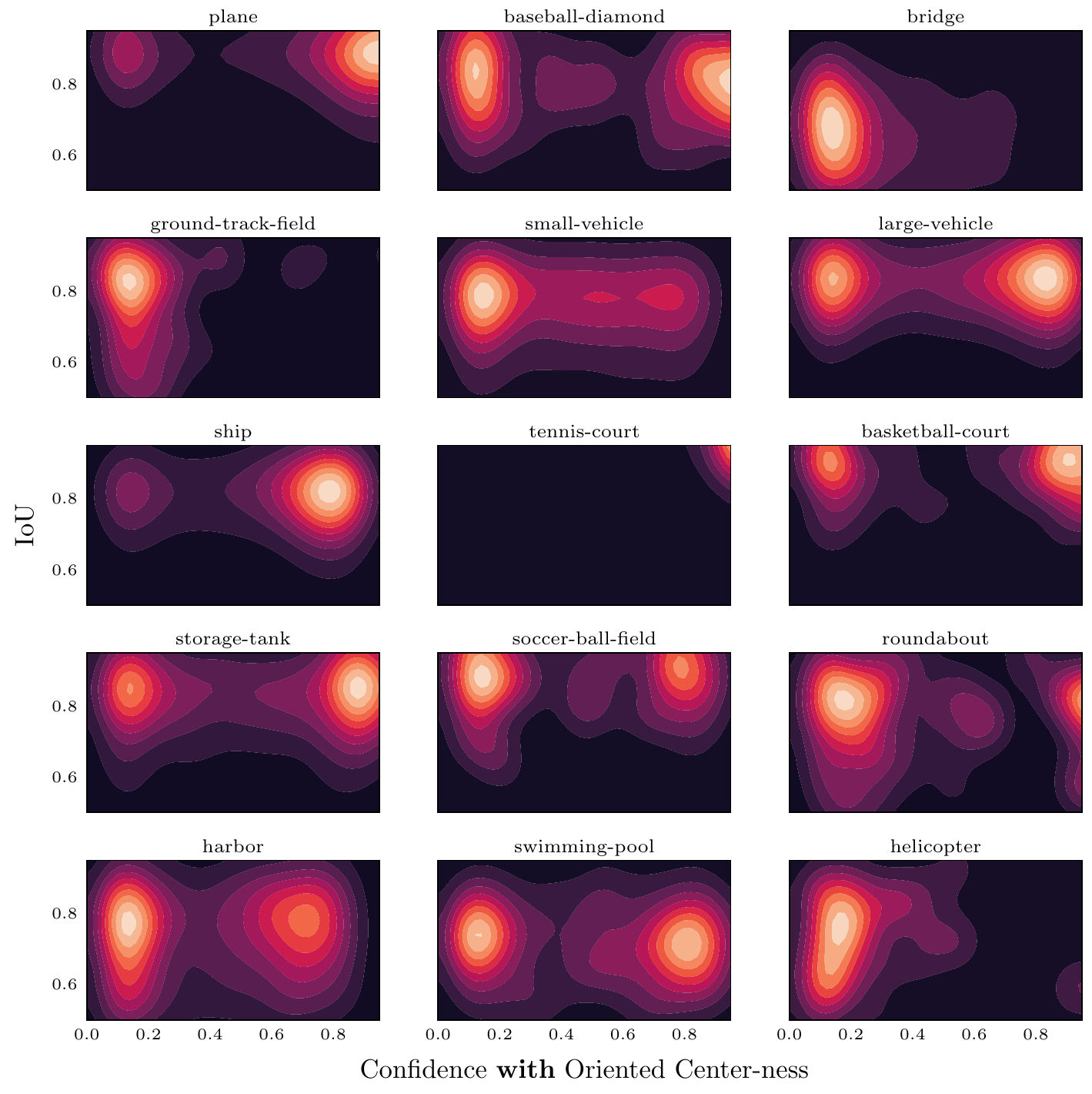}
  \caption{Heatmaps of classification confidences against IoU values for
    correctly detected oriented bounding boxes in the DOTA 1.0 validation set
    (true positives). Values are obtained from a model with oriented
    center-ness which results in higher classification confidence for
    correct detections and lower for the wrong ones. In this way, it is more likely that
    the wrong detections will be correctly removed in the post-processing steps.}
  \label{fig:exp:oriented-center-ness-classes-oriented}
\end{figure*}

We have collected the true positive detections as a heatmap to
better understand how the oriented center-ness improves the model
accuracy. The classification confidences for the 15 different classes of
the DOTA 1.0 validation set, obtained without oriented center-ness are depicted in
\figurename~\ref{fig:exp:oriented-center-ness-classes-baseline},
while the ones obtained with oriented center-ness are in
\figurename~\ref{fig:exp:oriented-center-ness-classes-oriented}.

The detection classification confidences are drawn against the Intersection over
Union (IoU) with the ground-truth bounding box. For each correctly
detected oriented bounding box in the validation set, a pair of
(\textit{confidence}, \textit{IoU}) is collected, where the IoU is computed
between the predicted bounding box and the ground-truth bounding box. The
heatmap brightness indicates the density of correctly detected objects in the
validation dataset. For a fair comparison, the values used for the heatmaps are the
confidences $p$ and not the
center-ness adjusted scores $s$. The inclusion of oriented center-ness during training as an
adjustment to the classification confidence results in detections of higher
confidences (and better IoU) already before scaling them with the oriented
center-ness value. This is appreciable in almost all the classes. In this way, the
detections with low center-ness values are correctly removed during
post-processing steps such as score thresholding and non-maximum suppression.
Moreover, the results obtained on classes such as ``helicopter'' indicate that future
works could explore how to combine two or more center-ness functions in a
fashion similar to ``Mixture of Experts''.

\section{DOTA 1.0 Evaluations per Class and Predictions Illustration}

\begin{table*}[!t]
  \caption{Test set accuracies on DOTA 1.0 (in mAP) of DAFNe compared with
    all other one-stage anchor-free oriented object detection models in the
    literature and with more complex two-stage and anchor-based models.
    Note that ``\mstrain'' indicates multi-scale
    training and testing, and ``\twostage'' indicates a two-stage
    model. Best results in bold, our approach is highlighted. Our model improves over the previous
    state-of-the-art in 7 out of 15 classes and it is either second best or
    close to the best scoring model for all other classes}
  \label{tab:exp:dota-class-maps}
  \begin{center}
    \begin{tabular}{lcccccccc}
      \toprule
      Method                                                                                         & PL             & BD             & BR             & GTF            & SV             & LV             & SH             & TC             \\
      \midrule
      \multicolumn{4}{l}{\textbf{Two-stage or                                         Anchor-based}} & & & & &                                  \\
      APE\mstrain\twostage\cite{zhu2020ape}                                                          & 89.96 & 83.62 & 53.42 & 76.03 & 74.01 & 77.16 & 79.45 & 90.83 \\
      OWSR\mstrain\twostage~\cite{Li_2019_CVPR_Workshops}                                            & \textbf{90.41} & 85.21 & 55.00 & 78.27 & 76.19 & 72.19 & 82.14 & 90.70\\
      KLD (RetinaNet)\mstrain~\cite{Yang2021LearningHB}                                              & 88.91 & 85.23 & 53.64 & 81.23 & 78.20 & 76.99 & 84.58 & 89.50 \\
      ReDet\mstrain\twostage~\cite{han2021ReDet}                                                     & 88.81 & 82.48 & 60.83 & 80.82 & 78.34 & \textbf{86.06} & 88.31 & 90.87 \\
      KLD (R\textsuperscript{3}Det)\mstrain\twostage~\cite{Yang2021LearningHB}                                & 89.92 & 85.13 & 59.19 & \textbf{81.33} & 78.82 & 84.38 & 87.50 & 89.80\\
      Oriented R-CNN\mstrain\twostage~\cite{Xie2021OrientedRF}                                       & 89.84 & \textbf{85.43} & \textbf{61.09} & 79.82 & \textbf{79.71} & 85.35 & \textbf{88.82} & \textbf{90.88} \\
      \midrulethin
      \multicolumn{4}{l}{\textbf{One-stage and                                        Anchor-free}}  & & & & &                                  \\
      IENet\mstrain~\cite{lin2019ienet}                                                              & 88.15          & 71.38          & 34.26          & 51.78          & 63.78          & 65.63          & 71.61          & 90.11          \\
      PIoU\mstrain~\cite{chen2020piou}                                          & 80.90          & 69.70          & 24.10          & 60.20          & 38.30          & 64.40          & 64.80          & 90.90          \\
      AxisLearning~\cite{xiao2020axis}                                                               & 79.53          & 77.15          & 38.59          & 61.15          & 67.53          & 70.49          & 76.30          & 89.66          \\
      AF-EMS~\cite{jia2021af-ems}                                                                    & \textbf{93.11} & 73.70          & 48.95          & 63.03 & 50.85          & 75.65          & \textbf{91.54} & \textbf{92.57} \\
      O\textsuperscript{2}-DNet~\cite{wei2020o2-dnet}                           & 89.31          & 82.14          & 47.33          & 61.21          & 71.32          & 74.03          & 78.62          & 90.76          \\
      P-RSDet\mstrain~\cite{zhou2020polar}                                                           & 88.58          & 77.84          & 50.44          & 69.29          & 71.10          & 75.79          & 78.66          & 90.88          \\
      CenterRot\mstrain~\cite{wang2021centerrot}                                                     & 89.74 & 83.57 & 49.53 & 66.45 & 77.07 & 80.57 & 86.97 & 90.75 \\
      MEAD~\cite{He2021MEADAM}                                                                       & 88.42 & 79.00 & 49.29 & 68.76 & 77.41 & 77.68 & 86.60 & 90.78 \\
      CFA\mstrain~\cite{Guo_2021_CVPR}                                                               & 89.26 & 81.72 & 51.81 & 67.17 & 79.99 & 78.25 & 84.46 & 90.77 \\
      BBAV\mstrain~\cite{Yi2021OrientedOD}                                                           & 88.63 & 84.06 & 52.13 & 69.56 & 78.26 & 80.40 & 88.06 & 90.87 \\
      AROA~\cite{he2021aroa}                                                                         & 88.33 & 82.73 & \textbf{56.02} & \textbf{71.58} & 72.98 & 77.59 & 78.29 & 88.63 \\
      \rowcolor{rowhighlightcolor} DAFNe\mstrain                                                     & 89.40          & \textbf{86.27} & 53.70 & 60.51          & \textbf{82.04} & \textbf{81.17} & 88.66          & 90.37          \\
      \\
                                                                                                     & BC             & ST             & SBF            & RA             & HA             & SP             & HC             & mAP           \\
      \midrule
      \multicolumn{4}{l}{\textbf{Two-stage or                                         Anchor-based}} & & & & &                                  \\
      APE\mstrain\twostage\cite{zhu2020ape}                                                          & 87.15 & 84.51 & 67.72 & 60.33 & 74.61 & 71.84 & 65.55 & 75.75 \\
      OWSR\mstrain\twostage~\cite{Li_2019_CVPR_Workshops}                                            & 87.22 & 86.87 & 66.62 & 68.43 & 75.43 & 72.70 & 57.99 & 76.36 \\
      KLD (RetinaNet)\mstrain~\cite{Yang2021LearningHB}                                              & 86.84 & 86.38 & 71.69 & 68.06 & 75.95 & 72.23 & 75.42 & 78.32 \\
      ReDet\mstrain\twostage~\cite{han2021ReDet}                                                     & 88.77 & 87.03 & 68.65 & 66.90 & \textbf{79.26} & \textbf{79.71} & 74.67 & 80.10 \\
      KLD (R\textsuperscript{3}Det)\mstrain\twostage~\cite{Yang2021LearningHB}                                & \textbf{87.33} & 87.00 & \textbf{72.57} & \textbf{71.35} & 77.12 & 79.34 & \textbf{78.68} & 80.63 \\
      Oriented R-CNN\mstrain\twostage~\cite{Xie2021OrientedRF}                                       & 86.68 & \textbf{87.73} & 72.21 & 70.80 & 82.42 & 78.18 & 74.11 & \textbf{80.87} \\
      \midrulethin
      \multicolumn{4}{l}{\textbf{One-stage and                                        Anchor-free}}  & & & & &                                  \\
      IENet\mstrain~\cite{lin2019ienet}                                                              & 71.07          & 73.63          & 37.62          & 41.52          & 48.07          & 60.53          & 49.53          & 61.24          \\
      PIoU\mstrain~\cite{chen2020piou}                                          & 77.20          & 70.40          & 46.50          & 37.10          & 57.10          & 61.90          & 64.00          & 60.50          \\
      AxisLearning~\cite{xiao2020axis}                                                               & 79.07          & 83.53          & 47.27          & 61.01          & 56.28          & 66.06          & 36.05          & 65.98          \\
      AF-EMS~\cite{jia2021af-ems}                                                                    & 59.29          & 73.17          & 49.87          & 69.00          & \textbf{81.79} & 66.99          & 58.33          & 69.86          \\
      O\textsuperscript{2}-DNet~\cite{wei2020o2-dnet}                           & 82.23          & 81.36          & 60.93 & 60.17          & 58.21          & 66.98          & 61.03          & 71.04          \\
      P-RSDet\mstrain~\cite{zhou2020polar}                                                           & 80.10          & 81.71          & 57.92          & 63.03          & 66.30          & 69.77          & 63.13          & 72.30          \\
      CenterRot~\cite{wang2021centerrot}                                                             & 81.50 & 84.05 & 54.14 & 64.14 & 74.22 & 72.77 & 54.56 & 74.00 \\
      MEAD~\cite{He2021MEADAM}                                                                       & 85.55 & 84.54 & 62.10 & 66.57 & 72.59 & 72.84 & 59.83 & 74.80 \\
      CFA\mstrain~\cite{Guo_2021_CVPR}                                                               & 83.40 & 85.54 & 54.86 & 67.75 & 73.04 & 70.24 & 64.96 & 75.05 \\
      BBAV\mstrain~\cite{Yi2021OrientedOD}                                                           & \textbf{87.23} & 86.39 & 56.11 & 65.62 & 67.10 & 72.08 & 63.96 & 75.36\\
      AROA~\cite{he2021aroa}                                                                         & 83.33 & 86.61 & \textbf{65.93} & 63.52 & 76.03 & 78.43 & 61.33 & 75.41 \\

      \rowcolor{rowhighlightcolor} DAFNe\mstrain                                                     & 83.81 & \textbf{87.27} & 53.93          & \textbf{69.38} & 75.61          & \textbf{81.26} & \textbf{70.86} & \textbf{76.95} \\
      \bottomrule
    \end{tabular}
  \end{center}

\end{table*}

\noindent DAFNe with oriented center-ness and center-to-corner prediction strategy 
sets the new state-of-the-art accuracy result on DOTA 1.0.
Table~\ref{tab:exp:dota-class-maps} shows the class-separated test accuracy (in
mAP) achieved by DAFNe and all its one-stage anchor-free competitors in the
literature, to the best of our knowledge, and by more complex two-stage and
anchor-based models. DAFNe improves accuracy over the
previous state-of-the-art in 7 out of 15 classes and it is either second best or
close to the best scoring model for all other classes. The acronyms refer to
the following classes: PL plane, BD baseball-diamond, BR bridge, GTF
ground-track-field, SH small-vehicle, LV large-vehicle, SH ship, TC
tennis-court, BC basketball-court, ST storage-tank, SBF soccer-ball-field, RA
roundabout, HA harbour, SP swimming-pool, HC helicopter.

\begin{figure*}
  \centering
  \includegraphics[width=1.0\linewidth]{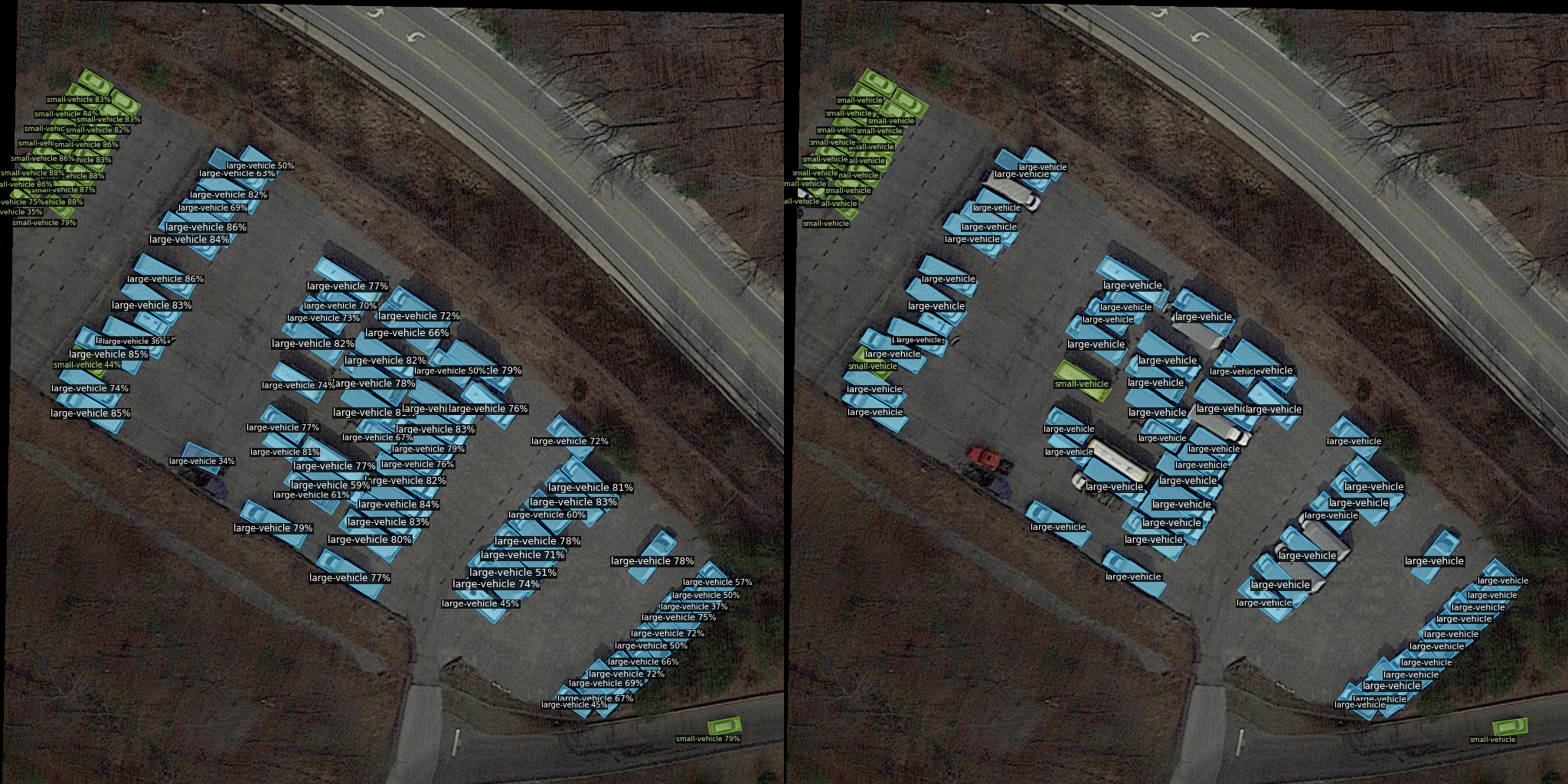}\\
  \includegraphics[width=1.0\linewidth]{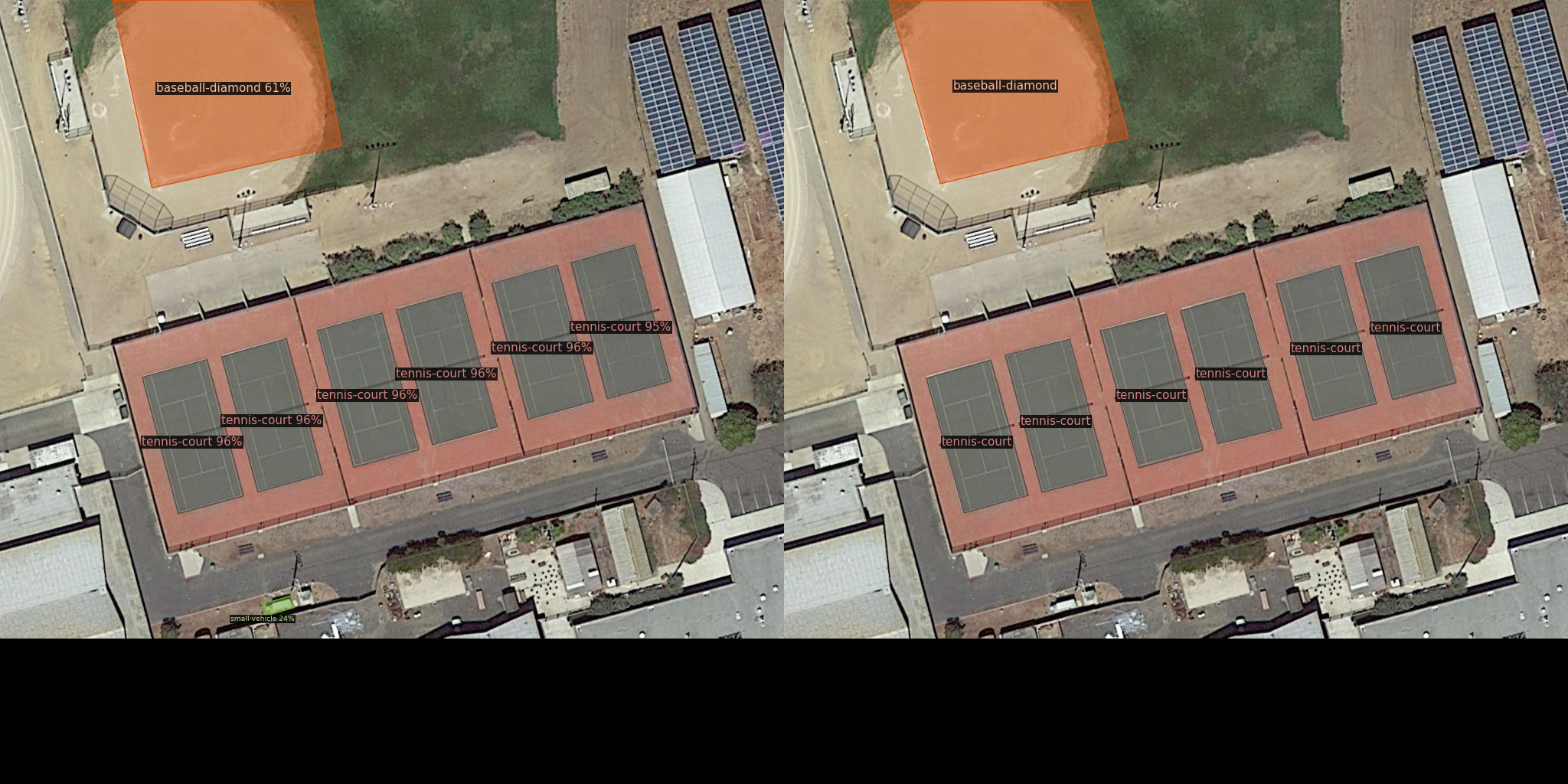}\\
  \caption{Accurate object predictions on some DOTA 1.0 validation set samples
    performed by DAFNe. In each couple of images, the left one shows object localizations and
    classifications with annotated confidence values, the right one shows
    the ground-truth objects positions and classes.}
  \label{fig:exp:baseline:preds-1}
\end{figure*}

\begin{figure*}
  \centering
  \includegraphics[width=1.0\linewidth]{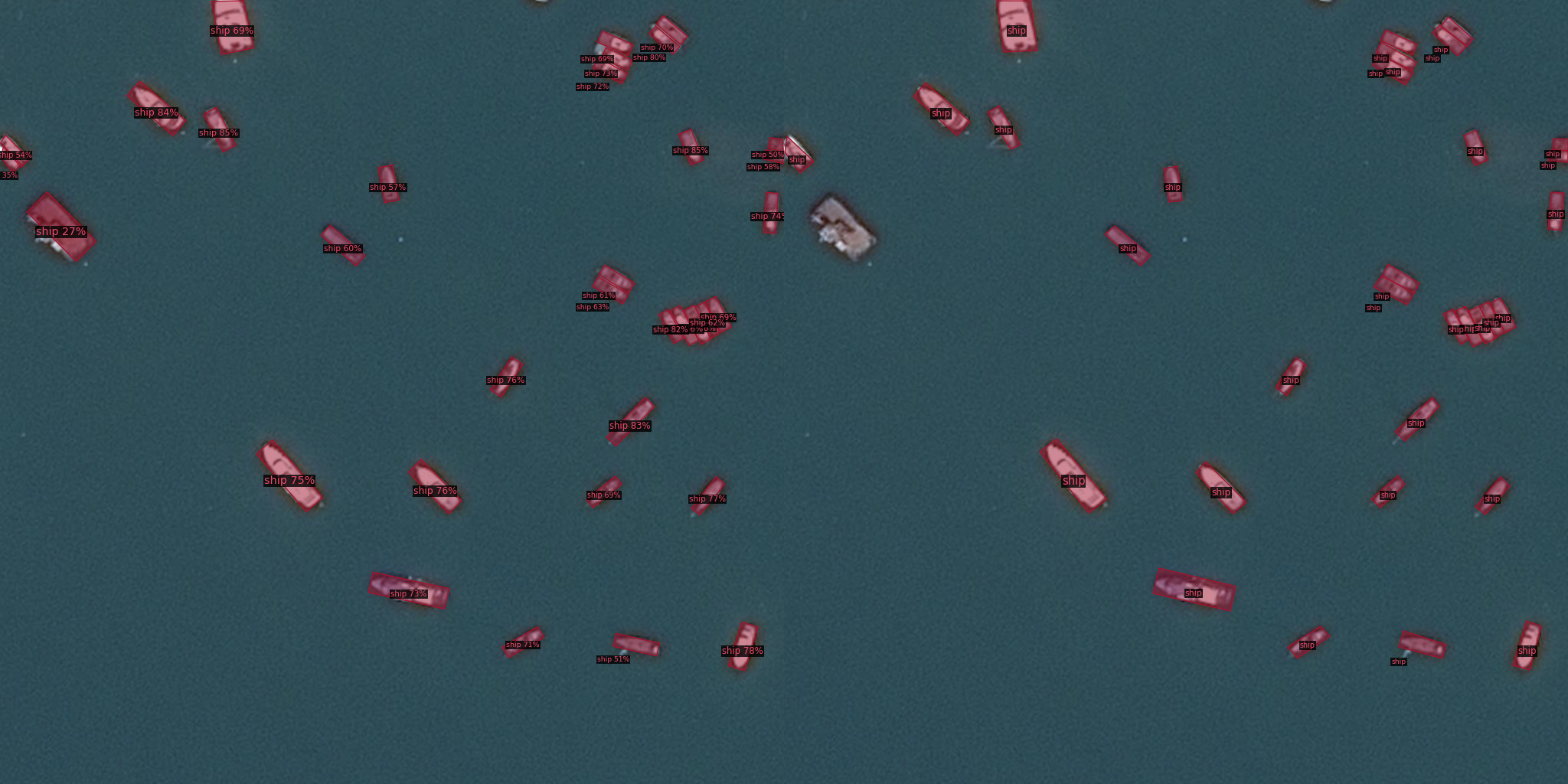}\\
  \includegraphics[width=1.0\linewidth]{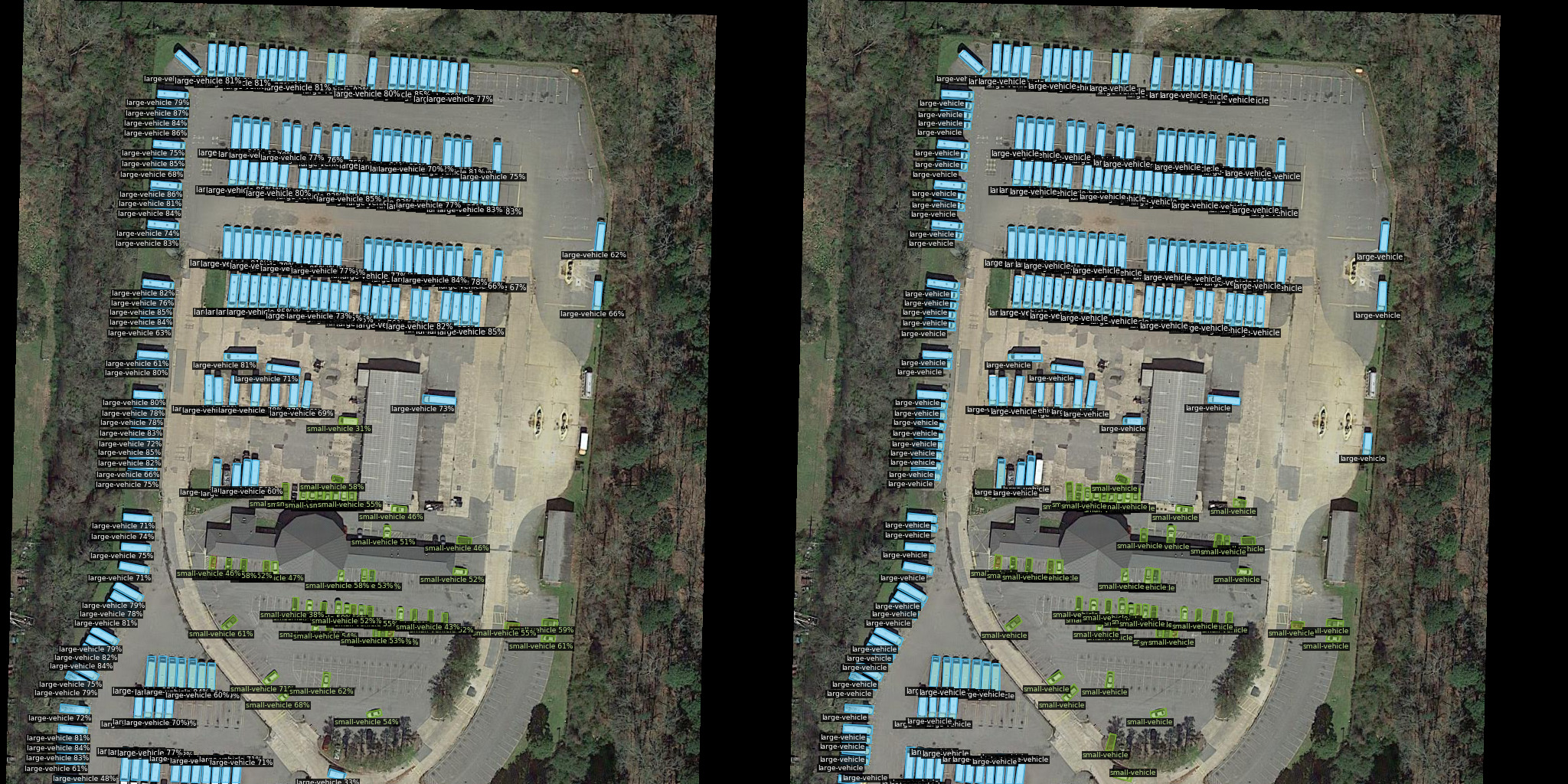}
  \caption{Accurate object predictions on some DOTA 1.0 validation set samples
    performed by DAFNe. In each couple of images, the left one shows object localizations and
    classifications with annotated confidence values, the right one shows
    the ground-truth objects positions and classes.}
  \label{fig:exp:baseline:preds-2}
\end{figure*}

\figurename~\ref{fig:exp:baseline:preds-1} and \figurename~\ref{fig:exp:baseline:preds-2} show accurate object predictions on
some DOTA 1.0 validation set samples performed by DAFNe. During our experiments,
we also have investigated how much DAFNe is sensitive to random seeds. We have
computed the DOTA 1.0 validation accuracy by using 10 different seeds obtaining
a negligible standard deviation of only 0.05\%, therefore, DAFNe is not
sensitive to random seeds.

\end{document}